\documentclass{article}
\usepackage[utf8]{inputenc}

\usepackage{geometry}
\usepackage{pdflscape}
\usepackage{subfig}
\usepackage{graphicx}
\usepackage{tikz}
\usetikzlibrary{positioning}

\usepackage{enumitem}
\usepackage{amsmath}
\usepackage{amsfonts}
\usepackage{hyperref}
\usepackage{xcolor}
\usepackage{comment}
\hypersetup{
    colorlinks = true,
    linkbordercolor = {white},
    linkcolor={gray},
    urlcolor={blue},
    citecolor= gray
}
\usepackage{commands}
\usepackage{color,soul}

\usepackage{authblk}

\title{Oralytics Reinforcement Learning Algorithm}
\author[1]{Anna L. Trella}
\author[1]{Kelly W. Zhang}
\author[2]{Stephanie M. Carpenter}
\author[3]{David Elashoff}
\author[4]{Zara M. Greer}
\author[5]{Inbal Nahum-Shani}
\author[6]{Dennis Ruenger}
\author[4]{Vivek Shetty}
\author[1]{Susan A. Murphy}
\affil[1]{Department of Computer Science, Harvard University}
\affil[2]{College of Health Solutions, Arizona State University}
\affil[3]{Division of General Internal Medicine and Health Services Research, Department of Biostatistics, and Department of Computational Medicine, University of California, Los Angeles}
\affil[4]{School of Dentistry, University of California, Los Angeles}
\affil[5]{Institute for Social Research, University of Michigan}
\affil[6]{Division of General Internal Medicine and Health Services Research, University of California, Los Angeles}
\begin{document}
\date{}
\maketitle

\begin{abstract}
    Dental disease is still one of the most common chronic diseases in the United States. While dental disease is preventable through healthy oral self-care behaviors (OSCB), this basic behavior is not consistently practiced. We have developed Oralytics, an online, reinforcement learning (RL) algorithm that optimizes the delivery of personalized intervention prompts to improve OSCB. In this paper, we offer a full overview of algorithm design decisions made using prior data, domain expertise, and experiments in a simulation test bed. The finalized RL algorithm was deployed in the Oralytics clinical trial, conducted from fall 2023 to summer 2024. 
\end{abstract}

\tableofcontents

%%%%%%%% PRELIMINARIES %%%%%%%
\section{Introduction}
Dental disease is one of the most common chronic diseases in the United States, particularly affecting disadvantaged communities. While scientific evidence indicates that healthy oral self-care behaviors (OSCB) (i.e., systematic, twice-a-day tooth brushing) prevent dental disease \cite{loe2000oral, attin2005tooth}, this basic behavior is not consistently practiced \cite{yaacob2014powered}. We have developed Oralytics, an online, reinforcement learning (RL) algorithm that optimizes the delivery of personalized intervention prompts to improve OSCB. These prompts, delivered via push notification from the Oralytics app, are designed to supplement clinician instruction and consist of engaging content tailored to participants, such as brushing feedback and motivational messages. This paper describes the methodology used to design and develop the online RL algorithm. To make quality design decisions, we leveraged prior data, domain expertise, and experiments in a simulation test bed. The online RL algorithm was deployed in the Oralytics clinical trial \cite{oralytics:clinicaltrial}. The main study of the clinical trial ran from fall 2023 to summer 2024. For more information on the trial design, please see \cite{nahum2024optimizing}. 
% A version of this document can be found in Appendix A of the Oralytics protocol paper \cite{nahum2024optimizing}.

\subsection{Preliminaries}
RL \cite{sutton2018reinforcement} is an area of machine learning where algorithms learn to select a sequence of decisions in an environment to maximize an outcome. In Oralytics, the RL algorithm optimizes the delivery of engagement prompts to participants to maximize their oral self-care behaviors (OSCB) (see Section~\ref{brush_quality_def}). We use $N$ to denote the number of participants participating in the main study (sample size). For the main study, the RL algorithm makes decisions for around $N = 70$ participants, where each participant is in the main study for 70 days. The RL algorithm makes decisions every day at decision points. For Oralytics, the decision points are right before the participant's morning and evening preferred brushing windows (Section~\ref{proximal_outcome_window}). To learn which decisions maximize OSCB for a participant, the RL algorithm updates with incoming participant data at update times. For Oralytics, the update times are every Sunday at 4:04 AM PST (Section~\ref{update_cadence}). Participants enter the study incrementally at an expected rate of 5 participants per 2 weeks (See Section~\ref{onboarding_proc} for exact details on the definition of a participant entering the RL part of the study).

% \subsection{Formulating the RL Problem}
To formulate the RL problem, we first define a state, action, and reward. For each component, we use subscript $i$ to denote the participant and subscript $t$ to denote the decision point. The state $S_{i, t}$ for participant $i$ at decision point $t$ describes the current context of the participant. We define algorithm state features as relevant features used by the algorithm that provide a summary of state $S_{i, t}$. For example, state features could be weather, location, or amount of dosage in the past week. See Section~\ref{state_features} for the state features for Oralytics. The action $A_{i, t}$ is the decision made by the algorithm for participant $i$ at decision point $t$, based on its policy. A policy is a function that takes in input state $S_{i, t}$ and outputs action $A_{i, t}$. For Oralytics, the action is a binary decision of whether to deliver an engagement prompt ($A_{i, t}=1$) or not ($A_{i, t}=0$), and the policy first calculates the randomization probability $\pi_{i, t}$ of selecting $A_{i, t}=1$ and then uses $\pi_{i, t}$ to sample $A_{i, t}$ from a Bernoulli distribution. That is, $A_{i, t}$ is micro-randomized at time $t$ with randomization probability  $\pi_{i, t}$. For Oralytics, the policy initially undergoes a warm-start period (Section~\ref{onboarding_proc}) where $\pi_{i, t}$ is calculated using the prior distribution (Section~\ref{fitting_prior}). The reward $R_{i, t}$ is a function of the proximal outcome,  designed to  reward the algorithm for selecting good actions for participants. In Oralytics, the reward is a function of OSCB (proximal health outcome concerning brushing quality) and a proxy for participant burden (cost of current treatment on the effectiveness of future treatment) \cite{trella2023reward}. See Section~\ref{reward_def} for more details on the reward designed for Oralytics.

We now describe how the RL algorithm makes decisions and learns throughout the study. The Oralytics algorithm is an \textit{online} RL algorithm that learns and updates throughout the study as new participant data accumulates. An online RL algorithm can be decomposed into two main parts: 1) a  statistical model of the participant (also called  the environment) and 2) a policy (defined above) informed by the participant  model. At decision point $t$, the algorithm observes participant $i$'s observed state $S_{i, t}$ and randomizes action $A_{i, t}$ with randomization probability $\pi_{i, t}$. After the RL algorithm selects $A_{i, t}$, the algorithm then observes the proximal outcome of OSCB $Q_{i, t}$ from the participant. At update times, the RL algorithm updates the model of the participant environment using a history of states, actions, and rewards up to the most recent decision point. 

\subsection{Available Data}
To inform our design decisions, we use data from ROBAS 2 \cite{info:doi/10.2196/17347} and ROBAS 3, previous dental health studies, and data from the pilot phase of the Oralytics study. All data we use is publicly available:
\begin{itemize}
    \item ROBAS 2: \href{https://github.com/ROBAS-UCLA/ROBAS.2/blob/master/inst/extdata/robas_2_data.csv}{here}
    \item ROBAS 3: \href{https://github.com/ROBAS-UCLA/ROBAS.3/blob/main/data/robas_3_data_complete.csv}{here}
    \item Oralytics Pilot: \href{https://github.com/StatisticalReinforcementLearningLab/oralytics_pilot_data}{here}
\end{itemize}

ROBAS 2 ran with 32 participants where each participant was in the study for 30 days. ROBAS 3 included 31 participants where each participant was in the study for 90 days. The pilot study included 9 participants and the study duration was 35 days. ROBAS 3 and the pilot study used the same sensory suite (i.e., Bluetooth toothbrush and sensory-collecting software). This suite is more sophisticated and was able to record brushing pressure to inform OSCB (a measure of brushing quality), while the suite for ROBAS 2 could only record brushing duration. A key distinction is that ROBAS 2 and ROBAS 3 had no interventions and the pilot study had interventions selected by a beta version of the RL algorithm. See Table~\ref{robas_3_stats} for additional statistics on the ROBAS 3 data set.

We use the ROBAS 2 data to inform certain design decisions (e.g., cost term parameters-- $b, a_1, a_2$, see Section~\ref{sec:cost_term}-- in reward definition). We use the ROBAS 3 data to design the simulation environment (i.e., fit participant environment models) (Section~\ref{app:sim_env}) and to inform additional design decisions (e.g., slope value of smoothing allocation function). Finally, we use the pilot study to develop the  prior distribution (Section~\ref{fitting_prior}) used by all algorithm candidates (all algorithms are Bayesian and thus require a prior distribution) and to set the participant app opening probability for the simulation environment (Section~\ref{sim_env_variants_open_app}).

One natural approach is to use pilot study data to create another simulation environment. However, the pilot study data is not suitable for creating a separate simulation environment because:

\begin{enumerate}
    \item \textbf{Insufficient Sample Size:} For the pilot study, we only have data on 9 participants.
    \item \textbf{Insufficient Duration of Pilot Study:} Each participant was in the pilot study for 35 days. In contrast, participants will be in the main study for 70 days. When running simulations, we evaluate our RL algorithm in a 70-day study. However, because we only have pilot participant data for 35 days,  the simulation environment built off of pilot data must extrapolate to the final 35 days (i.e., predict OSCB with day in study that was not in the training set).
\end{enumerate}

Both the sample size and duration of the pilot study (35 days)
means that we have data on only 70 decision
points for each of the 9 participants in the pilot study. The small amount of data  per participant  implies that the estimated  parameters in the models are very noisy. Therefore, we only use the pilot data to inform: 1) simulated app opening probability (Section~\ref{sim_env_variants_open_app})
and 2) the prior used by algorithm candidates that run in the simulation environment (Section~\ref{fitting_prior}). We continue to use the simulation environment built off ROBAS
3 data to inform design decisions and evaluate the final algorithm going into the Oralytics main study. 

Notice that although we do not have enough data to construct a new simulation
environment, we still use pilot
data to inform the design of the prior distribution for the algorithm. This is because we constructed the prior in collaboration with domain scientists (see Section~\ref{fitting_prior} for more information on constructing the prior).

\begin{table}[H]
\centering
\begin{tabular}{p{0.50\linewidth}p{0.175\linewidth}}
\toprule
Property & ROBAS 3 \\ 
\midrule
Num. Users & 31 \\
\% Sessions with No Brushing  & 53\% \\
Min. OSCB & 0 \\
Max. OSCB & 473 \\
SD. of OSCB & 74.908 \\
Mean of OSCB & 62.750 \\
Median of OSCB & 0.0 \\
\bottomrule
\end{tabular}
\caption{Descriptive statistics for OSCB in the ROBAS 3 data set}
\label{robas_3_stats}
\end{table}

\subsection{Code}
Code for running experiments to finalize design decisions for the RL algorithm can be found in a GitHub repository  \href{https://github.com/StatisticalReinforcementLearningLab/oralytics_algorithm_design}{here}.

%%%%%%%% ALGORITHM DESIGN DECISIONS %%%%%%%
\section{Algorithm Design Decisions}
\subsection{Overview}
%\sam{the overview is VERY helpful....} \alt{Thank you for acknowledging my contribution :)} 
The following list contains all decisions we made for the RL algorithm. Some decisions we  made in consideration with domain experts and other decisions define properties of algorithm candidates that we test in the simulation environment using the PCS framework for Reinforcement Learning \cite{a15080255}. Details on designing and constructing the simulation environment are in Section~\ref{app:sim_env}.

We categorize design decisions into two categories: 
\begin{enumerate}
    \item Fixed decisions that are informed using prior data and domain science
    \item Decisions made by experiments in the simulation test bed where each combination of decisions forms a unique algorithm candidate.
\end{enumerate}

\paragraph{Fixed Decisions:}
\begin{itemize}
    \item Participant Onboarding Procedure and Prior Sampling Period (Section~\ref{onboarding_proc})
    \item Proximal Outcome (Section~\ref{brush_quality_def})
    \item Duration of Proximal Outcome Window (Section~\ref{proximal_outcome_window})
    \item RL Framework (Section~\ref{rl_framework})
    \item Reward Approximating Function (Section~\ref{reward_approx_func})
    \item Algorithm State Features (Section~\ref{state_features})
    \item Forming the Prior (Section~\ref{fitting_prior})
    \item Dealing with app opening issue (Section~\ref{app_open_issue})
    \item Update Cadence (Section~\ref{update_cadence})
    \item Parameters of Smoothing Allocation Function (Section~\ref{smooth_allocation_func})
    \begin{itemize}
        \item Clipping Values
    \end{itemize}
    \item Reward Definition (Section~\ref{reward_def})
    \item Monitoring System (Section~\ref{monitoring})
\end{itemize}

\paragraph{Decisions made using experiments in the simulation testbed:}
\begin{itemize}
    \item Pooling Cluster Size (Section~\ref{pooling_cluster_size})
    \item Update Cadence (Section~\ref{update_cadence})
    \item Parameters of Smoothing Allocation Function (Section~\ref{smooth_allocation_func})
    \begin{itemize}
        \item Slope Value ($B = 0.515$ and $5.15$)
    \end{itemize}
    \item Hyperparameters for Reward Definition (Section~\ref{sec:cost_term})
\end{itemize}
\subsection{Participant Onboarding Procedure and Prior Sampling Period}
\label{onboarding_proc}
To warm start the RL algorithm and facilitate after-study analyses, the algorithm initially samples actions (whether to send an intervention prompt or not) according to the prior distribution (Section~\ref{fitting_prior}); the prior distribution represents our best guess at personalization based on the pilot participants’ data. This prior sampling period continues until the first update time (Section~\ref{update_cadence}) occurring after the 15th participant starts the study (participant starting the study defined below); the period after the first update time is called the \lq\lq RL period\rq\rq.
Notice that during the prior sampling period, the RL algorithm will still update the parameters in the participant model but will not use the updated model to select actions. Since the expected recruitment rate is 5 participants per 2 weeks or 2.5 participants per week, we expect around 5 weeks where the prior distribution is used to select actions.

We now define what it means for a participant to have started the study. First, we define a \textit{registered participant}. A participant is registered if the participant has completed the onboarding process with staff (i.e., received a unique study ID, downloaded the Oralytics app, and has a first successfully registered brushing session where data from the toothbrush has been successfully transmitted to the cloud). Next, we define the \textit{participant start date} as the first day the participant \textit{obtains the first schedule of actions} from the RL algorithm (more details about the schedule in Section~\ref{rl_algorithm_scheduler}). This happens when 1) the participant has successfully registered brushing sessions to the cloud, 2) a schedule has been formed for the participant, and 3) they have opened their app and the app has received the schedule. Once these conditions are met, we consider the participant as \textit{started the study}.
\subsection{Proximal Outcome}
\label{brush_quality_def}
The proximal outcome for Oralytics is oral self-care behaviors (OSCB) which is a measure of brushing quality. OSCB is the true behavioral outcome that the scientific team would like to improve for each participant to achieve long-term oral health. The choice of the proximal outcome is important because  it is used to evaluate RL algorithm candidates in the design phase with the simulation test bed. Notice the proximal outcome is different from the reward given to the algorithm to learn (Section~\ref{reward_def}). 

Denote the OSCB for participant $i$ and decision point $t$ to be $Q_{i, t}$. $Q_{i, t}$ is the proximal outcome, defined as $Q_{i, t} = \min(B_{i, t} - P_{i, t}, 180)$. $B_{i, t}$ is the participant's brushing duration in seconds and $P_{i, t}$ is the aggregated duration of over pressure in seconds. $Q_{i, t}$ is truncated by $180$ to avoid optimizing for over-brushing.

We also considered including other sensor information in our model of the OSCB, including zoned brushing duration, a measure of how evenly participants brush across the four zones (e.g. top-left quadrant, bottom-right quadrant, etc.). However, we did not end up including it in the OSBC because zoned brushing duration is not reliably obtainable as it requires Bluetooth connectivity and the participant to stand close enough to the docking station. Zoned brushing duration only appeared in about 82\% of brushing sessions in the pilot phase of the ROBAS 3 study.
\subsection{Duration of Proximal Outcome Window}
\label{proximal_outcome_window}
The time window over which we will collect the proximal outcome  is defined as the window of time following each decision point $t$ over which we record the proximal outcome $Q_{i, t}$ to action $A_{i, t}$. Recall that, as defined in Section~\ref{brush_quality_def}, $Q_{i, t} = \min(B_{i, t} - P_{i, t}, 180)$, where $B_{i, t}$ is the participant's brushing duration in seconds, and $P_{i, t}$ is the aggregated duration of over pressure in seconds.
Let $M_{i, d}, E_{i, d}$ be the morning and evening decision points for participant $i$ for day $d$. When a participant is onboarded, they are asked for their preferred morning and evening brushing times for weekdays and weekends such that the morning and evening decision points may differ between weekdays and weekends. Then the time window over which we collect the proximal outcome for participant $i$ is $[M_{i, d}, E_{i, d}]$ for the morning decision point and $[E_{i, d}, M_{i, d + 1}]$ for the evening decision point.

If a participant brushes more than once during a window, then the OSCB at the first brushing is the proximal outcome for that decision point. The algorithm will not use a state, action, OSCB $(S_{i, t}, A_{i, t}, Q_{i, t})$ data tuple when the algorithm forms the state features until that participant's window has concluded. That is, the algorithm  will process and format the most recent $(S_{i, t}, A_{i, t}, Q_{i, t})$ and construct the state features (such as the exponential average of OSCB over past 7 days) for each  participant as well as  construct the schedule of actions around 4 AM PST every night. By the above definition of the time windows, this means that data on the most recent evening OSCB will NOT be included in the state construction IF a participant's window has not ended.
\subsection{RL Framework}
\label{rl_framework}
We chose a Bayesian contextual bandit algorithm framework. 

\paragraph{1. Choice of using a Contextual Bandit Algorithm Framework:}
We understand that actions will likely affect a participant's future states and rewards (e.g., sending an engagement prompt the previous day may affect how responsive a participant is to an engagement prompt today). This suggests that an RL algorithm that models the participant's behavior by a Markov decision process (MDP) may be more suitable than just modeling the participant's reward as in a contextual bandit algorithm. 

However, the highly noisy environment and the limited data to learn from (140 decision points per participant total) make it difficult for the RL algorithm to accurately model state transitions. Due to estimation errors in the state transition model, the estimates of the delayed effects of actions used in MDP-based RL algorithms can often be highly noisy or inaccurate. This issue is exacerbated by our severely constrained state space (i.e., we have few features and the features we get are relatively noisy). As a result, an RL algorithm that fits a full MDP model may not learn much during the study, which could slow down personalization. 

To mitigate these issues, we use contextual bandit algorithms, which fit a simpler model of the participant. Using a lower discount factor (a form of regularization) has been shown to lead to selecting more effective actions than using the true discount factor, especially in data-scarce settings \cite{jiang2015dependence}. A contextual bandit algorithm can be interpreted as an extreme form of this regularization where the discount factor is zero. Finally, contextual bandits are the simplest algorithm for sequential decision-making and have been used to personalize digital interventions in a variety of areas
\cite{yom2017encouraging,DBLP:journals/corr/abs-1909-03539,figueroa2021adaptive,cai2021bandit}.  
Furthermore, we generalize the classical contextual bandit framework to consider the negative, delayed effects of actions by designing a reward for the RL algorithm \cite{trella2023reward} (Section~\ref{reward_def}). We design a reward that involves penalizing the proximal outcome by a cost term that can be viewed as a crude proxy for the delayed effect of actions in the Bellman equation in a MDP model of the participant. See Section~\ref{relation_to_mdp} for more details.

\paragraph{2. Choice of a Bayesian Framework:} We consider contextual bandit algorithms that use a Bayesian framework, specifically posterior (Thompson) sampling algorithms \cite{DBLP:journals/corr/0001RKO17}. Posterior sampling involves placing a prior distribution on the parameters in the model for the mean of the  reward conditional on state and action. This prior is used in the updates to form the  posterior distribution of these  parameters at each algorithm update time. To construct the prior distribution (Section~\ref{fitting_prior}) we use previous data and domain expertise. In addition, Thompson sampling algorithms are stochastic (action selections are randomized with probabilities depending on the posterior distribution), which helps the algorithm explore and learn better while facilitating causal inference analyses after the main study is completed.
\subsection{Reward Approximating Function }
\label{reward_approx_func}
An important decision in designing the contextual bandit algorithm is how to model the participant's reward.  The reward function is the conditional mean of the reward given state and action. We chose a Bayesian Linear Regression (BLR) to model the reward. The linear model for the reward is relatively simple, well-studied, and well-understood. In addition, BLR is easily interpretable by domain experts and allows them to critique and inform the model.

%%%%% NON-ACTION CENTERING VERSION %%%%%%
% The equation we use for the BLR reward approximating function is:
% \begin{equation}
% \label{eqn:blr}
%     R_{i, t} = m(S_{i, t})^T \alpha_0 + A_{i, t} f(S_{i, t})^T \beta + \epsilon_{i,t}
% \end{equation}
% where $\pi_{i,t}$ is the probability that the RL algorithm selects action $A_{i,t} = 1$ in state $S_{i,t}$ for participant $i$ at decision point $t$. $\epsilon_{i,t} \sim \mathcal{N}(0, \sigma_\epsilon^2)$ and there are priors on $\alpha_{0} \sim \mathcal{N}(\mu_{\alpha_0}, \Sigma_{\alpha_0})$, $\beta \sim \mathcal{N}(\mu_{\beta}, \Sigma_{\beta})$.

% \alt{Should we talk about why we didn't choose BLR with action centering? Due to the after-study inference considerations?}

\subsubsection{Bayesian Linear Regression with Action Centering}

Recall that our model for the reward is a Bayesian Linear Regression (BLR) model with action centering \cite{DBLP:journals/corr/abs-1909-03539}. Through experiments (Section~\ref{final_alg_decisions}), we 
made the final decision to use an algorithm that does full pooling (clustering with cluster size $N=70$). Full-pooling algorithms learn a single algorithm across all participants in the study and therefore share parameters $\alpha_0, \alpha_1, \beta$ across all participants.
The model of reward $R_{i, t}$ assumed by the algorithm is:
$$   R_{i, t} = f(S_{i, t})^T \alpha_0 +  A_{i, t}  f(S_{i, t})^T \beta + \epsilon_{i,t}$$
where
$\epsilon_{i,t} \sim \mathcal{N}(0, \sigma^2)$ (Section~\ref{fitting_prior}) and $f(S_{i, t})$ are algorithm state features (Section~\ref{state_features}). See (Section~\ref{reward_def}) for the definition of $R_{i,t}$. To enhance robustness to misspecification of the above model we use action centering \cite{DBLP:journals/corr/abs-1909-03539}. To use action centering, the algorithm learns an overparameterized version of the above model: 
\begin{equation}
\label{eqn:blr}
    R_{i, t} = f(S_{i, t})^T \alpha_0 + \pi_{i,t} f(S_{i, t})^T \alpha_1 + (A_{i, t} - \pi_{i, t}) f(S_{i, t})^T \beta + \epsilon_{i,t}
\end{equation}
where $\pi_{i,t}$ is the probability that the RL algorithm selects action $A_{i,t} = 1$ in state $S_{i,t}$ for participant $i$ at decision point $t$. We call the term $f(S_{i, t})^T \beta$ the advantage (i.e., advantage of selecting action 1 over action 0) and $f(S_{i, t})^T \alpha_0 + \pi_{i,t} f(S_{i, t})^T \alpha_1$ the baseline.
The priors are $\alpha_{0} \sim \mathcal{N}(\mu_{\alpha_0}, \Sigma_{\alpha_0})$, $\alpha_{1} \sim \mathcal{N}(\mu_{\beta}, \Sigma_{\beta})$, $\beta \sim \mathcal{N}(\mu_{\beta}, \Sigma_{\beta})$. We discuss how we set informative prior values for $\mu_{\alpha_0}, \Sigma_{\alpha_0}, \mu_{\beta}, \Sigma_{\beta}, \sigma^2$ in Section~\ref{fitting_prior}.

\subsubsection{Posterior Updating}
\label{posterio_update:blr}
During the update step, the reward approximating function will update the posterior with newly collected data. Since we chose a full pooling algorithm (Section~\ref{final_alg_decisions}), the algorithm will update the posterior using data shared between all participants in the study. 
%Additionally, we make $M$ draws of the parameters from the updated posterior and use them for all decision points until the next update time. 
Here are the procedures for how the Bayesian linear regression model performs posterior updating. 

Recall that in the main study, participants join the study incrementally.  
%from the main text that we simulate participants incrementally joining the study.
We use $t \in [1 \colon T]$ to index the $t^{\mathrm{th}}$ decision point for a given participant. We use $\tau$ to represent the $\tau^{\mathrm{th}}$ the week for update time  (the algorithm is only updated once a week at 4:04 AM PST every Sunday). The decision of update times is discussed in Section~\ref{update_cadence}. We use $\tau(i,t)$ to denote the function that takes in participant index $i$ and decision point $t$ and outputs the number of weeks since the RL part of the study started, up to and including the current week (which may not have completed).
%We use $\tau(i,t)$ to denote the function that takes in the participant index $i$ and the participant decision point $t$ and outputs the number of full weeks since the study started.
Suppose we are selecting actions for decision point $t$ for a participant $i$.
Let $\phi(S_{i,t}, A_{i,t}, \pi_{i,t}) = [f(S_{i, t})$, $\pi_{i, t}f(S_{i, t})$, $(A_{i, t} - \pi_{i, t})f(S_{i, t})]$ be the joint feature vector %\kwz{I suggest we just use $\phi(S_{i,t}, A_{i,t}, \pi_{i,t}) = [f(S_{i, t})$, $\pi_{i, t}f(S_{i, t})$, $(A_{i, t} - \pi_{i, t})f(S_{i, t})]$ instead since technically $\pi_{i,t}$ is not just a function of $S_{i,t}$ and $A_{i,t}$.} 
and $\theta = [\alpha_0, \alpha_1, \beta]$ be the joint weight vector. Notice that Equation~\ref{eqn:blr} can be vectorized in the form: $R_{i, t} = \phi(S_{i,t}, A_{i,t}, \pi_{i,t})^\top \theta + \epsilon$. Notice that $\theta$ is shared amongst participants because we are performing full pooling. Let $\Phi_{1:\tau(i,t) - 1} \in \mathbb{R}^{K \cdot (5 + 5 + 5)}$ be a matrix of all participants' data that have been collected in the study up to update-time $\tau(i,t)$, specifically, it is the matrix of all stacked vectors $\{\phi(S_{i,t}, A_{i,t}, \pi_{i,t}) \}$, where $K$ is the total number of participant decision points in the shared history (due to incremental recruitment this is not just $N \cdot (t-1)$).
Let $\mathbf{R}_{1:\tau(i,t)} \in \mathbb{R}^K$ be a vector of stacked rewards $\{ R_{i, t}\}$, a vector of all participants' rewards that have been collected in the study up to update-time $\tau$.

Recall that we have normal priors on $\theta$ where $\theta \sim \mathcal{N}(\mu^{\text{prior}}, \Sigma^{\text{prior}})$, $\mu^{\text{prior}} = [\mu_{\alpha_0}, \mu_{\beta}, \mu_{\beta}] \in \mathbb{R}^{5+5+5}$ and $\Sigma^{\text{prior}} = \text{diag}(\Sigma_{\alpha_0}, \Sigma_{\beta}, \Sigma_{\beta})$. At the update time $\tau$, $p(\theta | \HH_{1:n,\tau(i, t) - 1})$, the posterior distribution of the weights given current history $\HH_{1:n,\tau(i, t) - 1}$ for all participants who have entered the study, is conjugate and normal. 

\begin{equation*}
    \theta | \HH_{\tau} \sim \mathcal{N}(\mu_{\tau}^{\TN{post}}, \Sigma_{\tau}^{\TN{post}}) 
\end{equation*}
\begin{equation}
\label{post_var}
    \Sigma_{\tau}^{\TN{post}} = \bigg(\frac{1}{\sigma^2}\Phi_{1:\tau}^T \Phi_{1:\tau} + (\Sigma^{\TN{prior}})^{-1} \bigg)^{-1}
\end{equation}
\begin{equation}
\label{post_mean}
    \mu_{\tau}^{\TN{post}} = \Sigma_{\tau}^{\TN{post}} \bigg(\frac{1}{\sigma^2}\Phi_{1:\tau}^T \mathbf{R}_{1:\tau} + (\Sigma^{\TN{prior}})^{-1}\mu^{\TN{prior}} \bigg)
\end{equation}

% Note that we fit $\eta^2$ to the ROBAS 2 dataset and fixed it for all of our experiments. For the real study, we are considering assigning a conjugate prior to $\eta^2$ and updating it at update times.

\subsection{Algorithm State Features}
\label{state_features}

$S_{i,t} \in \mathbb{R}^d$ represents the $i$th participant's state at decision point $t$, where $d$ is the number of variables describing the participant's state. 

\subsubsection{Baseline and Advantage Features}
$f(S_{i,t}) \in \mathbb{R}^5$ denotes the features  used to model both the baseline reward function and the advantage. Notice that for Oralytics, the baseline and advantage features are the same (i.e., all $f(S_{i,t})$), but this is a design choice, and they do not have to be.

\paragraph{} These features are:
\begin{enumerate}
\label{alg_state_features}
    \item Time of Day (Morning/Evening) $\in \{0, 1\}$
    \item \label{alg_state:brushing} $\Bar{B}$: Exponential Average of OSCB Over Past 7 Days (Normalized) $\in \mathbb{R}$
    \item \label{alg_state:actions} $\Bar{A}$: Exponential Average of Engagement Prompts Sent Over Past 7 Days (Normalized)  $\in [-1, 1]$
    \item \label{alg_state:app} Prior Day App Engagement $\in \{0, 1\}$ (if the participant has the app open and in focus (i.e. not in the background))
    \item Intercept Term $\in \mathbb{R}$
\end{enumerate}
The normalization procedure for normalizing ``Exponential Average of Brushing Over Past 7 Days" is the same as the one described in Section~\ref{alg_stat_normalizations} for normalizing $\Bar{B}_{i,t}$ and $\Bar{A}_{i,t}$. We normalized state features to enhance the interpretability of the parameters in (Equation~\ref{eqn:blr}). 

Features \ref{alg_state:brushing} and \ref{alg_state:actions} are $\bar{B}_{i,t} = c_{\gamma}\sum_{j = 1}^{14} \gamma^{j-1} Q_{i, t - j}$ and $\bar{A}_{i,t} = c_{\gamma}\sum_{j = 1}^{14} \gamma^{j-1} A_{i, t - j}$ respectively, where $\gamma=13/14$. Recall that $Q_{i, t}$ is the proximal outcome of OSCB defined in Section~\ref{brush_quality_def} and $A_{i,t}$ is the treatment indicator. This is the same $\bar{B}_t, \bar{A}_t$ used in the cost term of the reward as described in Section~\ref{reward_def}. Notice that algorithm state features \ref{alg_state:brushing}, \ref{alg_state:actions} rely on participant data over the past 7 days. For the first 7 days of the participant in the study, we set these feature values to the average value of data collected so far. For example, if the participant's current day in study is 3, then feature \ref{alg_state:brushing} will be average brushing over days 1 and 2. Similarly, feature \ref{alg_state:actions} will be the average number of engagement prompts sent over days 1 and 2. In addition, recall that we do not use data for $\bar{B}_{i,t}, \bar{A}_{i,t}$ from a brushing window until that participant's window has ended (Section~\ref{proximal_outcome_window}). This means that data on the most recent evening OSCB will NOT be included in $\bar{B}_{i,t}$, and action selected for the most recent evening will also NOT be included in $\bar{A}_{i,t}$ if a participant's window has not ended.
% \sam{tell reader what data these exponential averages include.  That is review the text from the beginning on:  *****The algorithm will not use a state, action, OSCB $(S_{i, t}, A_{i, t}, Q_{i, t})$ data tuple until that participant's window has concluded. Note that constructing the schedule of actions for decision points happen at 2AM every night. A participant's brushing data is needed to  construct the state feature for exponential average of brushing over past 7 days. In accordance with the above definition of the time windows, this means that data on the most recent evening OSCB will NOT be included in  the state construction IF a participant's window has not ended (data tuple for current evening decision point will not be used to construct the state features). ****}
%\alt{Hello Susan, I put this reasoning above.}

\subsubsection{Normalization of State Features}
\label{alg_stat_normalizations}
We normalize features for numerical stability and to increase the interpretability of the parameters in (Equation~\ref{eqn:blr}) (i.e., all state features have meaning when equal to 0). $\bar{B}$ is designed to be between $[0, 180]$, but is now normalized to be between $[-1, 1]$. Similarly, $\bar{A}$ is designed to be between $[0, 1]$, but is now normalized to be between $[-1, 1]$.

\begin{equation}
    \label{v3_b_bar_norm}
    \text{Normalized} \; \bar{B} = \bigg(\bar{B} - \frac{181}{2}\bigg) / \frac{179}{2}
\end{equation}
\begin{equation}
    \label{v3_a_bar_norm}
    \text{Normalized} \; \bar{A} = 2 \cdot \bigg(\bar{A} - \frac{1}{2}\bigg)
\end{equation}

\subsubsection{Initial State Values}
Recall, that the RL algorithm uses the first brushing session to check if a participant is ``registered" (Section~\ref{onboarding_proc}). The first RL schedule does not use the first brushing session value to create state. The first two states used for the first schedule will use a 0 value for OSCB to start. The initial state that the algorithm uses to select actions for a participant's first day in the study is the following:
\\\\
\textit{Initial Morning State}: 
\begin{enumerate}
    \item Time of Day = 0
    \item $\Bar{B}$ = -1 (normalized average zero OSCB)
    \item $\Bar{A}$ = -1 (normalized value of no engagement prompts sent)
    \item Prior Day App Engagement = 0
    \item Intercept Term = 1
\end{enumerate}
\textit{Initial Evening State}: 
\begin{enumerate}
    \item Time of Day = 1
    \item $\Bar{B}$ = -1 (normalized average zero OSCB)
    \item $\Bar{A}$ = -1 (normalized value of no engagement prompts sent)
    \item Prior Day App Engagement = 0
    \item Intercept Term = 1
\end{enumerate}

Therefore, every participant who starts the study will have a ``full state" which is the initial state specified above.
\subsection{Prior}
\label{fitting_prior}
The prior distribution for the main study of Oralytics is constructed using the pilot study. 
%We use a different procedure than the one for fitting the prior used in the pilot study because the Oralytics pilot phase only consisted of usable data for 9 participants. We do not think we have sufficient data for asymptotic approximations in determining feature importance. Therefore 
We used the following procedure:
% note: each participant’s treatments are no longer independent because we have adaptively sampled data from running a pooling algorithm

\begin{enumerate}
    \item For each pilot participant $i \in [1:9]$, we fit a linear model with action-centering for the OSCB given state and action (using Equation~\ref{eqn:blr} but with $Q_{i, t}$ as the outcome and not $R_{i, t}$). 
    %\sam{ I realize that this might be the  reward with 100, 100 as tuning parameters.  We definitely need to tell reader if this is the case...} 
    %\alt{Hello Susan, we do not use the algorithm reward with the cost parameters, we fit the proximal outcome. I will change this to make it clear.}
    Notice that to prevent numerical instability, we fit each model using L2 regularization with $\lambda = 10^{-3}$. The linear model with action-centering contains 15 parameters. We report estimated parameter values for each dimension of the parameters across the 9 participants (Figure~\ref{fig:ac_model_params}) and report the estimated variance of the parameter estimators across 9 participants (Table~\ref{pilot_prior_stats}).
    \item We calculate standard effect sizes for each dimension of the parameters across the 9 participants (Section~\ref{calc_eff_sizes}) (Figure~\ref{fig:ac_model_standard_eff}). The computed standard effect sizes are used in discussion with domain scientists to determine feature importance and the finalized prior discussion (Section~\ref{finalized_prior}).
\end{enumerate}

\subsubsection{Finalized Prior}
\label{finalized_prior}
Recall that the Oralytics RL algorithm uses a Bayesian Linear Regression with Action-Centering reward approximating function (Equation~\ref{eqn:blr}). Therefore, we want to set priors on $\alpha_{0} \sim \mathcal{N}(\mu_{\alpha_0}, \Sigma_{\alpha_0})$, $\alpha_{1} \sim \mathcal{N}(\mu_{\beta}, \Sigma_{\beta})$, $\beta \sim \mathcal{N}(\mu_{\beta}, \Sigma_{\beta})$ and the noise variance $\sigma^2$.
Through experiments, we decided on a full-pooling algorithm (clustering with cluster size $N$). See Section~\ref{pooling_cluster_size} for a full discussion on why full-pooling was chosen. The algorithm uses data from  all participants to select each participant's treatment actions, thus using  the same parameters $\alpha_0, \alpha_1, \beta$ for all participants. To inform the development of the prior distribution for $\alpha_0, \alpha_1, \beta$, the scientific team, considering Figures~\ref{fig:ac_model_params}, \ref{fig:ac_model_standard_eff} and Table~\ref{pilot_prior_stats}  constructed the  prior in Table~\ref{tab:finalized_prior}.

\paragraph{Determining Feature Importance:} %Recall that we do not have sufficient data to use GEE analysis in determining feature importance. Instead, 
We calculated standardized effect sizes  for the parameters for each of  the 9 participants (Section~\ref{calc_eff_sizes}) and used these effect sizes to define feature importance. The overall guideline we followed was: if the absolute value of the mean standard effect size is greater than 0.15, then we consider that feature to be significant. However, we make an exception for the intercept in the advantage. Notice that the calculated standard effect size of the intercept in the advantage is negative and has an average magnitude greater than our threshold of 0.15.  Scientifically, the engagement prompts should either improve or not affect immediate OSCB (the intercept in the advantage should be non-negative), and thus our team decided to declare this intercept feature insignificant. After using these guidelines, we determined ``Time of Day", ``Average Past Dosage" and ``Intercept" to be significant for the baseline and ``App Engagement" to be significant for the advantage (Figure~\ref{fig:ac_model_standard_eff}).
% Intuition: We want to see for each feature, how the standardized effect sizes change. If the standardized effect sizes are close to 0, then that feature is not significant.

\paragraph{Setting Prior Means and Prior Variances} 
For significant features, we set the prior mean to the empirical mean parameter value for that feature across 9 participants (Figure~\ref{fig:ac_model_params}). For non-significant features, we set the prior mean to be 0. 
For significant features, we set the prior SD to the empirical SD for that feature across 9 participants. For non-significant parameters, we set the prior SD to the empirical SD divided by 2 (Table~\ref{pilot_prior_stats}). 
Notice that we are reducing the SD of the non-significant weights because we want to provide more shrinkage to the prior mean of 0. (i.e., more data is needed to overcome the prior). However, the reduction value of 2 was an arbitrary choice. 

\begin{table*}[!ht]
    \centering
    \begin{tabular}{c|c}
    \toprule
        Parameter & Oralytics Pilot \\
        \hline
        \midrule
        $\sigma^2$: noise variance & 3878 \\
        $\mu_{\alpha_0}$: prior mean of the baseline state features & $[18, 0, 30, 0, 73]^T$ \\
        $\Sigma_{\alpha_0}$: prior variance of the baseline state features & $\text{diag}(73^2, 25^2, 95^2, 27^2, 83^2)$ \\
        $\mu_{\beta}$: prior mean of the advantage state features & $[0, 0, 0, 53, 0]^T$ \\
        $\Sigma_{\beta}$: prior variance of the advantage state features & $\text{diag}(12^2, 33^2, 35^2, 56^2, 17^2)$ \\
    \end{tabular}
    \caption{\textbf{Finalized Prior Using Oralytics Pilot Data.} Values are rounded to the nearest integer. Recall that the ordering of the features is the same as described in Section~\ref{alg_state_features}: Time of Day, Exponential Average of Brushing Over Past 7 Days (Normalized), Exponential Average of Engagement Prompts Sent Over Past 7 Days, Prior Day App Engagement, Intercept Term.}
    \label{tab:finalized_prior}
\end{table*}

\paragraph{Setting the Noise Variance}
% \sam{we need to tell reader where how we got the value for the noise variance...}
To set the noise variance, we used the following procedure:
\begin{enumerate}
    \item We fit parameters to a linear model with action-centering, one per participant.
    \item We obtain the weights for each fitted model and calculate residuals (predicted proximal outcome of the model with the proximal outcome in the data).
    \item Noise variance $\sigma^2$ is set to the average empirical variance of the residuals across 9 participants.
\end{enumerate}

% \begin{table}[H]
%    \centering
% \begin{tabular}{ll}
% \toprule
%     & \textbf{Noise Variance $\sigma^2$} \\
%     \midrule
%     Action-Centering Model & 3878 \\
% \bottomrule
% \end{tabular}
% \caption{\textbf{Fitted Noise Variance.}
% Value is rounded to the nearest integer. To fit the noise variance, we did the following procedure:
% 1. We fit parameters to the model, one per participant.
% 2. We then obtain the weights for each fitted model and calculate residuals.
% 3. $\sigma^2$ is set to the average empirical variance of the residuals across 9 participants.
% }
% \label{fitted_noise_vars}
% \end{table}

% \begin{enumerate}
%     \item Time of Day (Morning/Evening) $\in \{0, 1\}$
%     \item $\Bar{B}$: Exponential Average of Brushing Over Past 7 Days (Normalized) $\in \mathbb{R}$
%     \item $\Bar{A}$: Exponential Average of Messages Sent Over Past 7 Days $\in [0, 1]$
%     \item Prior Day App Engagement $\in \{0, 1\}$ (if the participant has the app open and in focus: not in background)
%     \item Intercept Term $\in \mathbb{R}$
% \end{enumerate}

\subsubsection{Calculating Effect Sizes}
\label{calc_eff_sizes}
To inform the design of the final prior, we calculate standard effect sizes (Equation~\ref{standard_eff_eqn}) for each dimension of the parameters across the 9 participants (Figures~\ref{fig:ac_model_standard_eff}).

Let $\theta_i$ be the model parameters fit for participant $i$ using $T_i$ number of data points. For each participant $i$, we produce a standard noise standard deviation:
\begin{equation}
    \sigma_i = \sqrt{\frac{\sum_{t=1}^{T_i} (R_{i,t}- \bar{R}_i)^2}{T_i - 1}}
\end{equation}

where $|\theta_i|$ is the dimension of $\theta_i$ and $\bar{R_i} = \frac{1}{T_i} \sum_{t = 1}^{T_i} R_{i, t}$.
\\\\
The standard effect size for each dimension of the state features of a specific participant $i$ is:

\begin{equation}
    \label{standard_eff_eqn}
    \frac{\theta_i}{\sigma_i}
\end{equation}

\subsubsection{Supplementary Figures and Tables}
\begin{table}[H]
   \centering
\begin{tabular}{l|c}
\toprule
    & \textbf{Standard Deviation} \\
    \midrule
    Baseline \\ 
    \midrule
    Time of Day & 72.752 \\
    Average Past Brushing & 49.612 \\
    Average Past Dosage & 94.916 \\
    App Engagement & 54.391 \\
    Intercept & 82.510 \\
    \midrule
    Additional Baseline Due To Action-Centering \\ 
    \midrule
    Time of Day $\times \; \pi_{i, t}$ & 74.405 \\
    Average Past Brushing $\times \; \pi_{i, t}$ & 183.546 \\
    Average Past Dosage $\times \; \pi_{i, t}$ & 237.773 \\
    App Engagement $\times \; \pi_{i, t}$  & 143.272  \\
    $\pi_{i, t}$ & 129.336 \\
    \midrule
    Advantage \\ 
    \midrule
    Time of Day & 23.076 \\
    Average Past Brushing & 65.061 \\
    Average Past Dosage & 69.056 \\
    App Engagement & 56.367 \\
    Intercept & 34.604 \\
\bottomrule
\end{tabular}
\caption{\textbf{Standard Deviation of Fitted Parameters for the Action-Centering Model Across participants.}
Reported standard deviations are rounded to the nearest 3 decimal places. Notice that the estimated variability between participants may appear lower than expected. %true variability. 
This is because a pooling algorithm was run for the pilot study so the data across participants is not independent.
}
\label{pilot_prior_stats}
\end{table}

% \begin{enumerate}
%     \item We fit one linear regression model per participant.
%     \item We then obtain the weights for each fitted model and calculate residuals.
%     \item Compute the sum of squared residuals for each participant. Divide by T minus the dimension of the fitted parameters.
%     \item $\sigma_n^2$ is set to the average empirical variance of the residuals across 9 participants.
% \end{enumerate}

%%%%% BAR CHARTS %%%%%%
% \newgeometry{margin=1cm} % modify this if you need even more space
% \begin{landscape}
% \thispagestyle{empty}

% \begin{figure}[H]
%     \centering
%     \includegraphics[width=1.3\textwidth]
%     {figures/lin_model_params.pdf}
%     \caption{\textbf{Parameters of Pilot Data Fit To The Linear Model.} Values are rounded to the nearest 3 decimal places.}
%     \label{fig:lin_model_params}
% \end{figure}
% \end{landscape}
% \restoregeometry

\newgeometry{margin=1cm} % modify this if you need even more space
\begin{landscape}
\thispagestyle{empty}
\begin{figure}[H]
    \centering
    \includegraphics[width=1.3\textwidth]
    {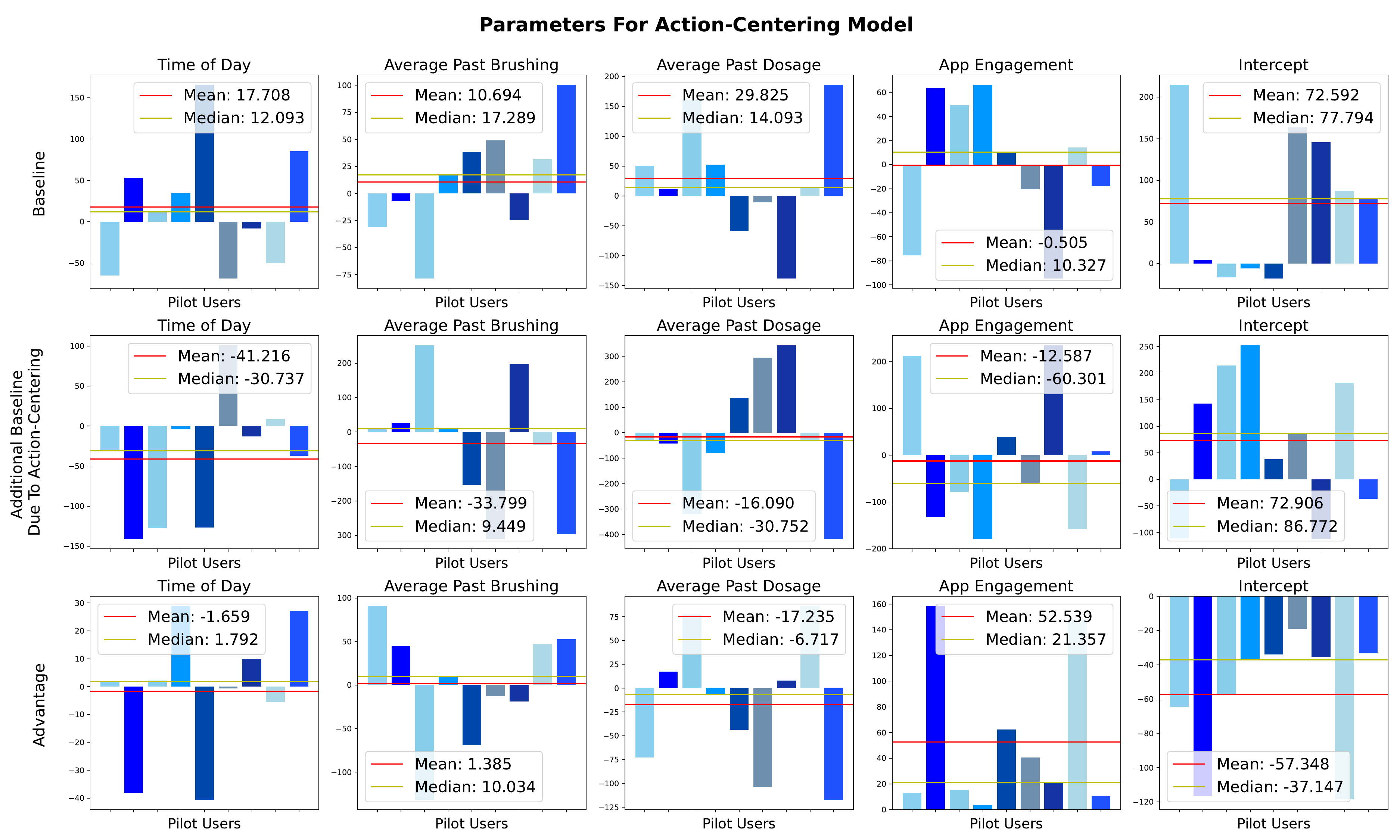}
    \caption{\textbf{Parameters of Pilot Data Fit To The Action-Centering Model.}}
    \label{fig:ac_model_params}
\end{figure}

\end{landscape}
\restoregeometry

%%%%% EFFECT SIZES %%%%%%
\newgeometry{margin=1cm} % modify this if you need even more space
\begin{landscape}
\thispagestyle{empty}
\begin{figure}[H]
    \centering
    \includegraphics[width=1.3\textwidth]
    {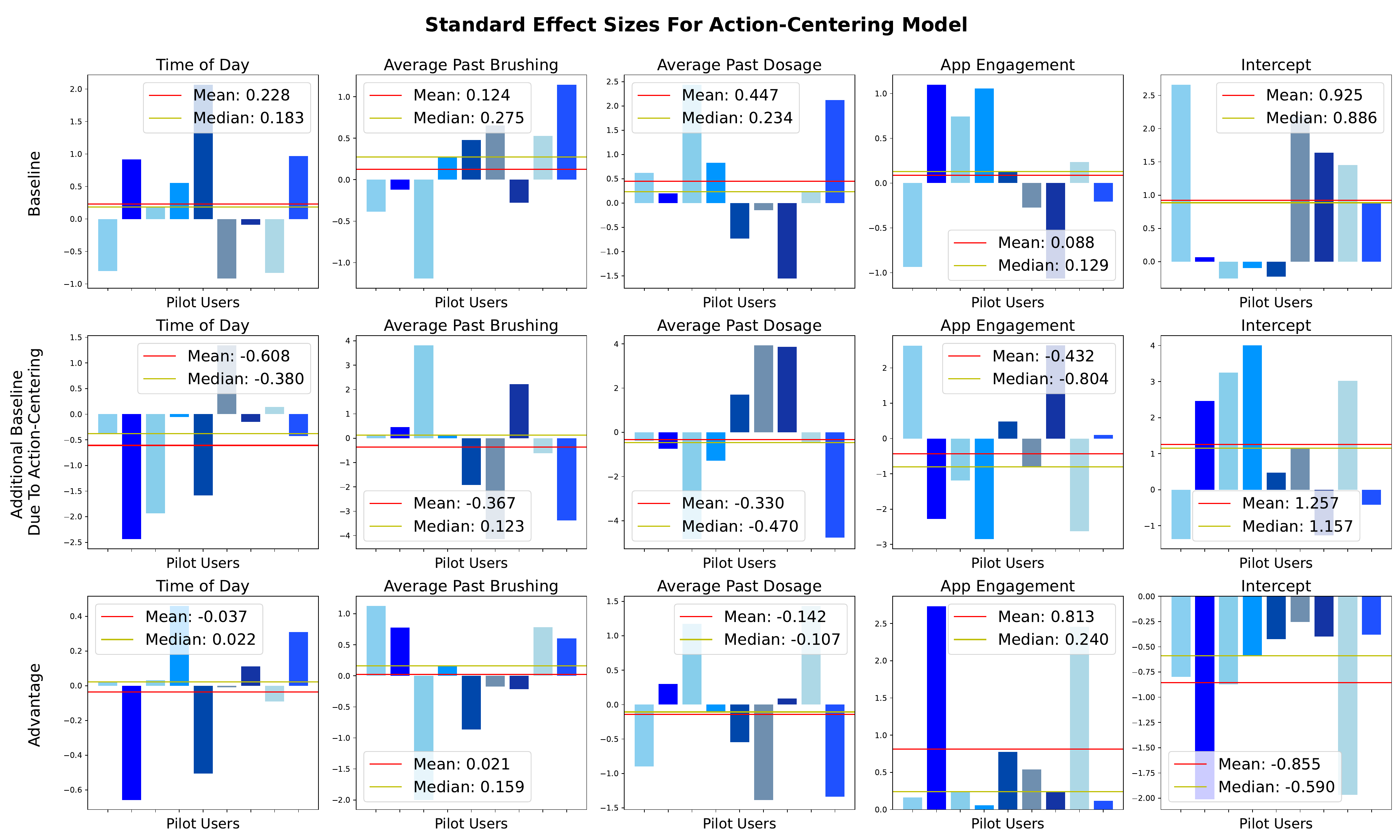}
    \caption{\textbf{Standard Effect Sizes From The Action-Centering Model.}}
    \label{fig:ac_model_standard_eff}
\end{figure}

\end{landscape}
\restoregeometry

\subsection{Dealing with App Opening Issue}
\label{app_open_issue}
We have a cloud-based system for the Oralytics RL algorithm. The cloud-based system consists of 1) a sensory collecting device (i.e. Bluetooth toothbrush), 2) a cloud server (Bayesian RL algorithm calculates posterior distribution for parameters in reward model and constructs $\pi_{i,t}$), and 3) device that relays the current action to the participant (i.e. mobile phone app). Due to limitations in native mobile app development and a computational need to run the RL algorithm on the cloud (and not locally on the mobile phone): \textit{the algorithm can only relay the most recent action when the participant opens the app.} (Note: the algorithm obtains the current state and reward regardless of the participant opening the app because the sensory collecting device directly communicates with the cloud server.) Requiring the participant to open their app multiple times a day is unrealistic. Therefore, we want to modify the RL algorithm to deal with cases where the participant does not open the app. In addition to modifying the RL and designing the intervention prompts to encourage participants to open the app, we developed a protocol for staff to encourage participants to open the app.

\subsubsection{Scheduling Solution With Modified RL Algorithm}
\label{rl_algorithm_scheduler}
Ideally, we want each action selection to be based on the most current data. However, in case the participant does not open the app we have the RL algorithm provide a full 70-day (the entire length of the study) schedule of actions starting with the current decision point $t$. The RL algorithm will return a matrix in $\mathbb{R}^{70 x 2}$ where each row represents the action selected for the participant $i$'s morning and evening decision points for that day. For decision points $j = t, t + 1$ (current day), the algorithm uses the current state $S_{i, t}, S_{i, t + 1}$ (See Section~\ref{state_features}) as input to $\pi_{i,t}$  to select actions. For decision points $j = t + 2, t + 3,..., t + 26, t + 27$ (within first 2 weeks from $t$) the algorithm will use a modified feature space as input to $\pi_{i,t}$ (See Section~\ref{app:modified_rl_features}). For all decision points after $t + 27$, the algorithm selects actions with a fixed probability $0.5$.

We chose the scheduling solution as a way to deal with the app opening issue because it is simple to implement and monitor. We chose to schedule a full 70-day schedule because prompts will still be delivered to a participant even in the worst case where a participant does not open their app ever again during the study. Having decision points $j = t + 2, t + 3,..., t + 26, t + 27$ use the modified feature space for action selection while additional decision points after that use randomization with fixed probability $0.5$ was a design decision made in consideration with domain experts. This decision was made because domain experts believed the modified state space was still valuable to use within 2 weeks. After 2 weeks, the modified state space may be too stale to be used for selecting actions. 

\subsubsection{Modified Feature Space of the RL Algorithm}
\label{app:modified_rl_features}
The modified state space is:
\begin{enumerate}
    \item \label{modstate:tod} Time of Day (Morning/Evening) $\in \{0, 1\}$
    % [0, 180]
    \item \label{modstate:brushing} \textit{Most Recent} Exponential Average of OSCB Over Past 7 Days (Normalized) $\in \mathbb{R}$
    \item \label{modstate:a_bar} Exponential Average of Engagement Prompts Sent Over Past 7 Days $\in [0, 1]$
    \item \label{modstate:app} \textit{Best Guess of} Prior Day App Engagement (Opened App / Not Opened App) $= 0$
    \item \label{modstate:bias} Bias / Intercept Term $\in \mathbb{R}$
\end{enumerate}

Notice that we know the values of state features (Section~\ref{state_features}) except for features \ref{alg_state:brushing} ($\bar{B}$) and \ref{alg_state:app} (prior day app engagement). Features \ref{modstate:tod} and \ref{modstate:bias} we know deterministically. We also know feature \ref{modstate:a_bar} by looking at previous sent actions and actions we have selected in the schedule. Since we do not know $\bar{B}$ ahead of time, we impute this feature with the most recent value of $\bar{B}_{i, t}$ known when the schedule is formed. Namely, the $\bar{B}$ value for decision points $j = t + 2, t + 3,..., t + 26, t + 27$ uses the same feature $\bar{B}$ value as decision points $t, t+1$.  For the prior day app engagement feature, we impute the value to 0, because our best guess is that the participant is not getting a fresh schedule because they did not open the app.

\subsubsection{Staff Protocol}
\label{staff_protocol}
We also employ a protocol to encourage participants to open the app:

\begin{enumerate}
    % \item The intervention prompts are designed to encourage participants to open the app
    \item Research staff will send participants a text message if they do not open the app for 5 days and a call if they don't open the app 5 days after getting a text message.
    \item There is a 30-day touch point meeting where the participant is paid for attending and staff can ask the participant to open the app
\end{enumerate}
\subsection{Pooling Cluster Size}
\label{pooling_cluster_size}
Clustering involves grouping $K$ participants together and pooling all $K$ participants' data together for the RL algorithm (i.e., the algorithm uses the history of all participants $i$ in the same cluster to update, and the same algorithm is used to select actions for all participants in the cluster). We consider clustering because clustering-based algorithms have been empirically shown to perform well when participants within a cluster are similar \cite{zhu2018group,tomkins2021intelligentpooling}. We also believe that clustering will facilitate learning within environments that have noisy within-participant rewards \cite{deshmukh2017multi,vaswani2017horde}. We decided on a full-pooling algorithm after running experiments (Section~\ref{final_alg_decisions}).

Before finalizing the full-pooling decision, we conducted experiments with cluster sizes $K$ = 1 (no pooling) and $K = N = 70$ (full pooling). There is one RL algorithm instantiation per cluster (no data shared across clusters). There is a trade-off between no pooling and full pooling. No pooling may learn a policy more specific to the participant later on in the study but may not learn as well earlier in the study when there is not a lot of data for that participant. Full pooling may learn well earlier in the study because it can take advantage of all participants' data but may not personalize as well as a no-pooling algorithm, especially if participants are heterogeneous. We found through our experiments that full-pooling algorithms outperformed no-pooling algorithms across all variants of the simulation environment (Section~\ref{final_alg_decisions}).

% For our experiments, we draw $N = 70$ simulated participants (the expected sample size for the Oralytics study) with replacement  \sam{from the ROBAS 3 participants?  REader is confused.  Need more explanation!  Might be better to put this language about the simulation later where the simulation is discussed!} \alt{you are correct Susan, moving this to the experiments section instead of keeping this here.} and cluster these participants at random (every possible cluster is equally likely). For each trial, we redraw participants and cluster assignments.

\subsection{Update Cadence}
\label{update_cadence}
Update cadence refers to the cadence or time at which the RL algorithm updates the posterior distribution. Namely, at update times, the RL algorithm updates the posterior distribution of the parameter in the reward approximating function (Section~\ref{posterio_update:blr}) using all the data that is available up to that time. We decided on a weekly update cadence after running experiments (Section~\ref{final_alg_decisions}). The RL algorithm updates every Sunday at 4:04 AM PST.

Before finalizing the decision on a weekly update cadence, we ran experiments considering a daily and weekly cadence. We are interested in both cadences because we believe a more frequent update cadence will enable the RL algorithm to  learn faster and therefore select better actions that yield better OSCB outcomes for participants. However, we found in our simulations that the difference between daily and weekly updates was relatively small (see Section \ref{experiments}); one reason behind this could be the app opening issue which prevents the algorithm from relaying the most recent action information when the participant does not open the app (see Section \ref{app_open_issue}). For this reason and to reduce the computational burden for computing standard errors for the primary analysis \cite{zhang2022statistical}, we decided on a weekly update cadence.
%However, we are considered that too many update cadences make after-study analyses difficult. \kwz{KELLY TODO: could you please explain more why we wanted a weekly update cadence?} 
% \alt{Anna, I made edits above. Please check to make sure they are accurate.}
%\alt{looks great Kelly! Thank you!}
\subsection{Smoothing Allocation Function}
\label{smooth_allocation_func}
The RL algorithm is a modified posterior sampling algorithm called the smooth posterior sampling algorithm. Recall in Section~\ref{reward_approx_func}, our model for the reward is a Bayesian linear regression model with action centering:
\begin{equation*}
    R_{i, t} = m(S_{i, t})^T \alpha_0 + \pi_{i,t} f(S_{i, t})^T \alpha_1 + (A_{i, t} - \pi_{i, t}) f(S_{i, t})^T \beta + \epsilon_{i,t}
\end{equation*}
where $\pi_{i,t}$ is the probability that the RL algorithm selects action $A_{i,t} = 1$ in state $S_{i,t}$ for participant $i$ at decision point $t$. $\epsilon_{i,t} \sim \mathcal{N}(0, \sigma^2)$ and there are priors on $\alpha_{0} \sim \mathcal{N}(\mu_{\alpha_0}, \Sigma_{\alpha_0})$, $\alpha_{1} \sim \mathcal{N}(\mu_{\beta}, \Sigma_{\beta})$, $\beta \sim \mathcal{N}(\mu_{\beta} \Sigma_{\beta})$.

Recall that the RL algorithm micro-randomizes the actions using $\pi_{i,t}=\PP \left( \action{i}{t} = 1 \big| \HH_{1: n, \tau(i,t) - 1}, \state{i}{t}=s \right)$.   The RL algorithm sets  
\begin{equation}
    \begin{aligned}
    \pi_{i,t}
    = \E_{\tilde{\beta} \sim \N(\mu_{\tau(i,t) - 1}^{\TN{post}}, \Sigma_{\tau(i,t) - 1}^{\TN{post}} )} \left[ \rho(s^\top \tilde{\beta}) \big| \HH_{1:n, \tau(i,t) - 1}, \state{i}{t}=s \right]
    \end{aligned}
\end{equation}
Recall above we use $\tau(i,t)$ to denote the function that takes in participant index $i$ and decision point $t$ and outputs the number of full weeks since the main study started, up to and including the current week (which may not have completed).
% \sam{Anna, please check above as the posterior sampling only uses the posterior from the most recent update time.  In display (8) should we replace $\tau(i,t)$by $\tau(i,t)-1$?}  \kwz{What I'm saying in Appendix B is: $\tau(i,t)$ is used to denote the function that takes in the participant index $i$ and the participant decision point $t$ and outputs the number of weeks since the MRT study started, up to and including the current week (which may not have completed). \alt{Anna agrees with Kelly's definition.}
% $$\pi_{i,t} \triangleq \PP \left( A_{i,t} = 1 \big| \HH_{1:n,\tau(i,t)-1}, S_{i,t} \right) \\
%     = \E_{(\alpha_0, \alpha_1, \beta) \sim \N ( \bar{\mu}_t, \bar{\Sigma}_t )} \left[ \rho \big( f(S_{i,t})^\top \beta \big) \big| \HH_{1:n,\tau(i,t)-1}, S_{i,t} \right].$$
% }
Notice that the last expectation above is only over the draw of $\beta$ from the posterior distribution parameterized by $\mu_{\tau(i,t) - 1}^{\TN{post}}$ and $\Sigma_{\tau(i,t) - 1}^{\TN{post}}$ (see displays \eqref{post_mean} and \eqref{post_var} for their definitions).

In classical posterior sampling, the posterior sampling algorithm uses an indicator function:
\begin{equation}
    \rho(x) = \II(x > 0)
\end{equation}
If the indicator function above is used, the posterior sampling algorithm sets randomization probabilities to the posterior probability that the treatment effect is positive.
% \begin{equation}
%     \begin{aligned}
%     \pi_{i,t}
%     = \E_{\tilde{\beta} \sim \N(\mu_{\mathrm{post}}, \Sigma_{\mathrm{post}})} \left[ \II(s^\top \tilde{\beta} > 0 ) ~ \big| ~ \history{1:n}{t-1}, \state{i}{t}=s \right].
%     \end{aligned}
% \end{equation}

\begin{figure}[H]
\centering
\begin{tikzpicture}
  \node (img)  {\includegraphics[width=8cm]{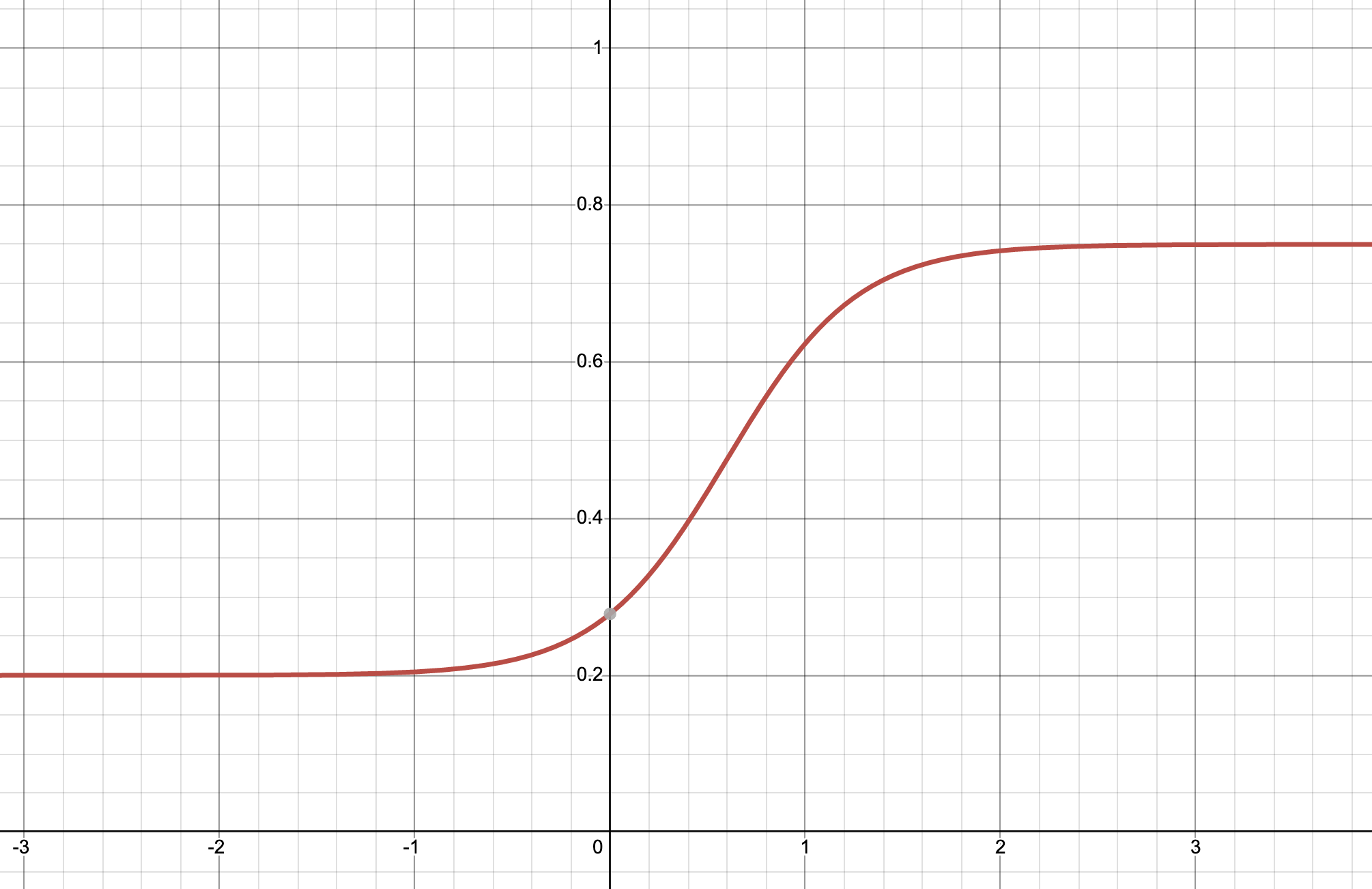}};
\node[below=of img, node distance=0cm, yshift=1cm,font=\color{black}] {$x$};
\node[left=of img, node distance=0cm, rotate=90, anchor=center,yshift=-0.7cm,font=\color{black}] {$\rho(x)$};
\end{tikzpicture}
\caption{
Generalized logistic function with $L_{\min}=0.2$ (lower clipping), $L_{\max}=0.8$ (upper clipping), $c = 5$ (shift to right), $b = 20$. We show the function with $b = 20$ instead of the chosen $b=\frac{20}{\sigma_{\text{rvv}}} = 0.515$ to help behavioral scientists interpret the target probability of sending an engagement prompt given the treatment effect standardized by the residual standard deviation. %Recall that $\sigma_{\text{rrv}}$ is the standard deviation of the reward residual variance of the ROBAS 3 data set: $\sigma_{\text{rrv}} = 38.83$. 
%\kwz{todo: label the axes; add a note for the motivation for adding the residual variance; explaination under the figure - interpolating between flat and steep}
}
\label{smooth_func_plot}
\end{figure}

In order to facilitate after-study inference with a full-pooling algorithm \cite{zhang2022statistical},
we replace the indicator function with a ``smooth" $\rho$. Using a smooth function $\rho$ ensures the policies formed by the algorithm concentrate. Concentration enhances the replicability of the randomization probabilities if the study is repeated. 
Without concentration, the randomization probabilities might fluctuate greatly between repetitions of the same study. For more discussion of this see \cite{deshpande2018accurate,zhang2020inference,NEURIPS2021_49ef08ad,zhang2022statistical}. For Oralytics, we chose $\rho$ to be a \href{https://en.wikipedia.org/wiki/Generalised_logistic_function}{Generalized logistic function}, which allows us to change the asymptotes and centering of a standard logistic function:
\begin{equation}
    \rho(x) = L_{\min} + \frac{ L_{\max} - L_{\min} }{ \big[ 1 + c \exp(-b x) \big]^k}
\end{equation}
See Figure~\ref{smooth_func_plot} for a plot of the smoothing allocation function. Above, 
\begin{itemize} 
    \item $L_{\min}=0.2$ and $L_{\max}=0.8$. $L_{\min}, L_{\max}$ are the upper and lower asymptotes. We chose these values to be the upper and lower clipping values but they do not have to be.
    \item $c=3$. Larger values of $c > 0$ shift the value of $\rho(0)$ to the right.  This choice implies that $\rho(0)=0.3$.
    \item $k=1$. Larger values of $k > 0$ make the asymptote towards the upper clipping less steep and the asymptote towards the lower clipping more steep
    \item $b = \frac{20}{ \sigma_{\text{rrv}}} = 0.515$. Larger values of $b > 0$ makes the slope of the curve more ``steep''. We choose $b$ to be the desired steepness divided by the standard deviation of the reward residual variance from ROBAS 3, $\sigma_{\text{rrv}} = 38.83$. We finalized the  steepness, $b$, using simulations (See Section~\ref{experiments}). The flatter the slope, the closer the action-selection probabilities are to $\rho(0)=0.3$ randomization. The steeper the slope, the more the action-selection probabilities approach an indicator function %(I can trust the data more. The more confident we are.
\end{itemize}

% \kwz{KELLY TODO: Hey Kelly, I am trying to update the graph below using desmos and parameters specified in the caption, but I can't recreate the graph without it looking really weird. Can you help me?} \alt{Hey Anna, I'm not sure how you wanted to update the graph, but in the graph below, the plot is for $b=20$, not for $b=0.515$.}

% ref: https://tex.stackexchange.com/questions/316481/adding-axes-labels-to-latex-figures

\subsection{Reward Definition}
\label{reward_def}
Recall in Section~\ref{brush_quality_def} we defined the behavioral health proximal outcome. Although the goal is to still maximize the outcome $Q_{i, t}$, we have flexibility in defining the reward given to the algorithm %(i.e., the reward $R_{i, t}$ does not necessarily have to be directly equal to $Q_{i, t}$) 
to be a function of $Q_{i, t}$ in order to improve the algorithm's learning. In simulation we compare the Oralytics RL algorithm variants in terms of their ability to maximize each participant's total OSCB,  $\sum_{t=1}^{T} Q_{i, t}$, where $Q_{i,t}$ is a non-negative measure of OSCB observed after each decision point (two times a day) (definition of $Q_{i, t}$ in Section~\ref{brush_quality_def}).  

We now discuss the design of the reward that will be used by the RL algorithm. Throughout, we are interested in optimizing for OSCB $Q_{i, t}$ and refer to $R_{i,t}$, the reward used by the RL algorithm, as the \textit{surrogate reward}. The RL uses the surrogate rewards $R_{i, t}$ to update the posterior distribution of the parameters in Bayesian Linear Regression. Specifically, let $R_{i, t} \in \mathbb{R}$ denote the surrogate reward for the $i$th participant at decision point $t$:
\begin{equation}
\label{reward}
        R_{i, t} := Q_{i, t} - C_{i, t}
\end{equation}

\subsubsection{Cost Term $C_{i, t}$}
\label{sec:cost_term}
The cost term is designed to allow the RL algorithm to optimize for immediate healthy brushing behavior, while simultaneously considering the delayed effect of the current engagement prompt on the effectiveness of future interventions. The cost term can be interpreted as a function (with parameters $\xi_1, \xi_2$) which takes in current state $S_{i, t}$ and action $A_{i, t}$ and returns the delayed negative effect of currently sending an engagement prompt (Section~\ref{relation_to_mdp}).

Recall in Section~\ref{state_features} we defined $\bar{B}_{i,t} := c_{\gamma}\sum_{j = 1}^{14} \gamma^{j-1} Q_{i, t - j}$ and $\bar{A}_{i,t} := c_{\gamma}\sum_{j = 1}^{14} \gamma^{j-1} A_{i, t - j}$. Note that in the algorithm state features, $\bar{B}, \bar{A}$ are normalized, but in the cost term below they are not. We set $\gamma = \frac{13}{14}$ to represent looking back 14 decision point points and scale each sum by constant $c_{\gamma}=\frac{1-\gamma}{1-\gamma^{14}}$ so that the weights sum to 1. Notice that our choice of $\gamma$ and the scaling constant means $0\le\bar{B}_{i,t}\le 180$ and $0\le \bar{A}_{i,t}\le 1$.
$\bar{B}_{i,t}$ captures the participant's exponentially discounted OSCB in the past week. $\bar{A}_{i,t}$ captures the number of interventions that were sent over the past week. 
Both terms are exponentially discounted because we expect that interventions sent and participant brushing in the near past will be more predictive of the delayed impact of the actions (i.e., affecting a participant's responsivity to future actions) than those in the further past.
% Billie Note for Susan: I know you are interested in habit formation and I think there is a connection here. If the participant is a high performer during a week in which they received many prompts, it would make sense to penalize intervention delivery because its possible that the person has formed a habit and it is time to tone down intervention delivery to facilitate more independence; here, the disadvantage of sending a prompt is not related to habituation or burden, but to a missed opportunity to support independence and autonomy  

We define the cost of sending an engagement prompt (i.e. captures participant burden in sending a prompt) as:
\begin{equation}
\label{cost_term}
C_{i, t} := 
\begin{cases}
\xi_1 \mathbb{I}[\bar{B}_{i, t} > b] \mathbb{I}[\bar{A}_{i, t} > a_1] & \\
\hspace{10mm} + \xi_2 \mathbb{I}[\bar{A}_{i, t} > a_2]  & \smash{\raisebox{1.6ex}{if $A_{i, t} = 1$}} \\
0 & \hspace{-0mm} \mathrm{if~} A_{i, t} = 0
\end{cases}
\end{equation}

% explaining the terms
Notice that the algorithm only incurs a cost if the current action is to send an intervention, i.e., $A_{i, t} = 1$.  The first term $\xi_1 \mathbb{I}[\bar{B}_{i, t} > b] \mathbb{I}[\bar{A}_{i, t} > a_1]$ encapsulates the belief that if a high-performing participant was sent too many engagement prompts within the past week, then we want to penalize the reward. The second term $\xi_2 \mathbb{I}[\bar{A}_{i, t} > a_2]$ encapsulates the belief that regardless of participant performance, if they received too many engagement prompts within the past week, then we also want to penalize the reward. $b, a_1, a_2$ are chosen by domain experts. Notice that $a_1 < a_2$ because we believe a high-performing participant will have a lower threshold of being burdened by an engagement prompt. The scientific team decided to set the following values:

\begin{itemize}
    \item $b=111$, is set to the 50th-percentile of participant brushing durations in ROBAS 2.
    %, where brushing durations are truncated to 120 seconds if they exceed 120. 
    % \sam{the truncation does not impact the value of the 50th percentile so   I commented it out...}
    \item $a_1 = 0.5$, represents a rough approximation of the participant getting an engagement prompt 50\% of the time (rough approximation because we are using an exponential average mean) 
    \item $a_2 = 0.8$, represents a rough approximation of the participant getting an engagement prompt 80\% of the time (rough approximation because we are using an exponential average mean) 
\end{itemize}
$\xi_1, \xi_2$ are non-negative hyperparameters that we tune (Section~\ref{algorithm_experiments}). %\sam{tell reader where to find this tuning discussion.}

\subsubsection{Reward Design To Improve Learning}
\label{relation_to_mdp}
By designing the reward in this way, we can approximate the delayed, negative effects of actions, the same way an MDP-based algorithm would, but allows us to maintain the bandit algorithm framework \cite{trella2023reward}.
% Our design and definition of the reward is the same as the reward described in. 

Based on domain knowledge, we believe that sending an engagement prompt at a decision point $t$ can only have a non-negative effect on the participant's immediate OSCB, $Q_{i,t}$. However, sending too many engagement prompts can risk habituation or may burden the participant thus affecting participant responsivity to future engagement prompts, i.e., affecting $Q_{i,t+1}, Q_{i,t+2}, \dots, Q_{i,T}$. 
Therefore, to anticipate these negative delayed effects of sending an engagement prompt, we reduce the algorithm's reward when negative delayed effects are likely to occur.  $C_{i, t}$ provides this reduction as including $C_{i, t}$ in the algorithm's reward will provide a signal that sending an engagement prompt ($A_{i, t} = 1$) may negatively affect future states. This signal is needed because we are using a contextual bandit-type RL algorithm that does not explicitly model the delayed effects of actions.

$C_{i, t}$ can be viewed as a crude proxy for the delayed effect of actions in the Bellman equation in a MDP model for the participant. Recall that according to the Bellman equation, it is optimal to select action $1$ over action $0$ if the immediate expected reward received from action $1$ over action $0$ exceeds the difference in optimal ``future values'' of selecting action $0$ over action $1$; specifically action $1$ and action $0$ can differ in ``future value'' due to the probability that each action will lead to a favorable or less favorable next state. 
Mathematically, this difference in future value is $\mathbb{E}[V^*(S_{i, t + 1}) | S_{i, t}, A_{i, t} = 0] - \mathbb{E}[V^*(S_{i, t + 1}) | S_{i, t}, A_{i, t} = 1]$ where $V^*$ is the optimal value function in a MDP setting. 
Note that in a pure contextual bandit setting the difference in future values of two actions is always zero, i.e., it is assumed that there are no delayed effects of actions on future states. By including a cost on selecting action $1$, we can move from an always zero model of delayed effects of sending an engagement prompt (selecting action $1$) used by the contextual bandit algorithm, to a more realistic setting in which there is some non-negative delayed effect of sending an engagement prompt, captured by our cost term  $C_{i,t}$.
\subsection{Monitoring System}
\label{monitoring}
Autonomously monitoring an RL algorithm during the study is important for detecting, alerting, and preventing errors. Errors could arise during the study that lead to critical situations (e.g., participants receiving too many or no engagement prompts). %The likelihood of these errors occurring increases because the RL algorithm learns and updates throughout the study.
Therefore, we have designed a quality monitoring system to prevent incidents and help detect problems as soon as they occur. This allows the research team to identify, triage, and solve issues, minimizing time and negative impact on participants and the study system.

The monitoring system involves a list of issues, alarms that check for these issues, and an automatic emailing system that emails the research team when an alarm is triggered.
% An autonomous monitoring system is especially important as it can take the burden off of staff from constantly checking clinician dashboards or verifying the system is properly working. An undetected issue or an issue that takes too long to fix can significantly jeopardize study validity, the effectiveness of treatment, and study results.

\subsubsection{Issues}
The following section details issues that  the system monitors. These issues are:
\begin{itemize}
    \item \textbf{Insufficient Dosage:} if a participant received less than the minimum amount of messages over a week
    \item \textbf{Excessive Dosage:} if a participant received the maximum amount of messages over a week
    \item \textbf{Impacting Validity of Data}: errors that compromise the scientific usability of the data for stakeholders or other research teams (e.g., issues that corrupt the data or prevent the system from saving necessary data needed for after-study analyses)
    \item \textbf{Impacting Prompt Randomization:} errors that impact the algorithm forming a schedule of prompts for the participant (e.g., issues obtaining state features for the participant).
    \item \textbf{Impacting Algorithm Updating:} errors that impact the algorithm updating the parameters in the model for the distribution of the reward (e.g., issues obtaining the most recent history of states, actions, and rewards).
\end{itemize}

\section{Algorithm Design Experiments}
\label{algorithm_experiments}
%\sam{Anna, please check that following sentence is accurate.} 
Our simulation environment creates a simulated participant  based on each of the participants in  ROBAS 3 and simulates participant app opening behavior based on participants' app engagement data from the Oralytics pilot study. See Section~\ref{app:sim_env} for details on the simulation environment. To run the final set of experiments, we first specify RL algorithm candidates that we want to evaluate (see next section).  Next, we run those candidates in each variant of the simulation environment (Table~\ref{tab:env_variants}).

\subsection{Algorithm Candidates}
These are the decisions made using experiments with the simulation environment. For each decision, we consider the following possibilities:
\begin{itemize}
    \item Pooling Cluster Size (No pooling $K = 1$ vs. Full pooling $K=N$) (Section~\ref{pooling_cluster_size})
    \item Update Cadence (Daily vs. Weekly) (Section~\ref{update_cadence})
    \item Parameters of Smoothing Allocation Function (Section~\ref{smooth_allocation_func})
    \begin{itemize}
        \item Slope Value ($B = 0.515$ and $5.15$)
    \end{itemize}
    \item Hyperparameters for Reward Definition ($\xi_1, \xi_2 \in [0, 180]$) (Note: We first run simulations of all other algorithm dimensions specified above with $\xi_1, \xi_2 = [100, 100]$. After all design decision have been made, we fix all other design decisions and tune $\xi_1, \xi_2$ using the procedure described in \cite{trella2023reward}.)
\end{itemize}

\subsection{Simulation Environment Variants}
\label{sim_env_vars}
%\sam{Need to clarify for reader. I realise that some of this will be clarified in the simulation environments section  but reader will be so confused that the reader is likely to give up.    I think reader is going to be pretty confused about how a simulated participant is generated for each environment variant.  Would be good to explain this somewhere....  For example a simulated participant's rewards are not generated using action centering and also by now reader has forgotten where to find the values of the weights/parameters used in the simulation... } 
To finalize the RL algorithm for the main study, we run algorithm candidates in twelve simulation environment variants (Table~\ref{tab:env_variants}).

The variant axes are:
\begin{itemize}
    \item \textbf{Environment Feature Space:} We consider two environments, stationary vs. non-stationary (see more information in Section~\ref{baseline_features}).
    \item \textbf{Level of Decline in Participant Responsivity:} We consider three levels of participant robustness to declining responsivity $E = \{0, 0.5, 0.8\}$ (see more information in Section~\ref{sim_env_variants_delayed_effs} on how we formulate $E$ to reduce treatment effect sizes). This declining responsivity can be due to habituation, burden, or independence of the participant from needing a prompt to brush well.
    \item \textbf{Population-Level Effect Sizes:} We consider a reasonable small effect size (shrinkage value $\zeta=\frac{1}{8}$) and a less small effect size (shrinkage value $\zeta=\frac{1}{4}$) (see more information in Section~\ref{treatment_eff_sizes}).
\end{itemize}
% See Section~\ref{baseline_features} for the differences between the stationary and non-stationary environments. 
% \sam{would be good to include the formula for how we use $E$--participant robustness to reduce effect sizes.}
% See Section~\ref{sim_env_variants_delayed_effs} for a discussion of how we use $E$ (participant robustness) to reduce effect sizes. See Section~\ref{treatment_eff_sizes} on how we simulated treatment effects with the two effect size scales. 

\begin{table}[H]
%    \centering
\begin{tabular}{lll}
\toprule
    \textbf{STAT\_LOW\_R:} Stationary Base Model, $E=0$  \\
    \midrule
    \textbf{NON\_STAT\_LOW\_R} Non-Stationary Base Model, $E=0$ \\
    \midrule
    \textbf{STAT\_MED\_R:} Stationary Base Model, $E=0.5$ \\
    \midrule
    \textbf{NON\_STAT\_MED\_R} Non-Stationary Base Model, $E=0.5$ \\
    \midrule
    \textbf{STAT\_HIGH\_R} Stationary Base Model $E=0.8$ \\
    \midrule
    \textbf{NON\_STAT\_HIGH\_R} Non-Stationary Base Model, $E=0.8$ \\ 
\bottomrule
\end{tabular}
\caption{{\textbf{Six Environment Variants For Each Effect Size Scale}}.
There are 12 total environment variants. For each effect size scale (small $\frac{1}{4}$ and smaller $\frac{1}{8}$), we consider two environment base models (stationary and non-stationary) and three levels of robustness to declining responsivity by sent engagement prompts $E$ (low $E = 0$, medium $E = 0.5$, and high $E = 0.8$). $E=0$ means participants are not robust and $E=0.8$ means they are fairly robust.}
%Shrinkage values shrink the participant-specific effect size if a participant receives too many engagement prompts. Low values of $E$ means that participants are more susceptible to habituation than higher values of $E$.
\label{tab:env_variants}
\end{table}
\subsection{Experiments}
\label{experiments}
We evaluate the RL algorithm candidates in each of the environment variants. For our experiments, we simulate the app opening issue (Section~\ref{app_open_issue}) we will face in the real study. We simulate an incremental recruitment rate of five participants entering the study every 2 weeks. To simulate a study, we sample $N = 70$ participants (approximately the expected sample size for the Oralytics study) with replacement from the 31 simulated participants developed using ROBAS 3 data. We then cluster sampled participants by their entry date into the simulated study. Further, all algorithm candidates have a study-level prior sampling period (Section~\ref{onboarding_proc}). For full-pooling algorithm candidates, the algorithm performs action selection using the prior until the next update time after the 15th participant enters the study. For no-pooling algorithm candidates, there is one algorithm per participant and the algorithm performs action selection using the prior during the first week for each participant. After the prior sampling period, the algorithm uses the posterior to perform action selection.

All algorithm candidates use a contextual bandit framework with Thompson sampling, a Bayesian Linear Regression with action centering reward approximating function, $[0.2, 0.8]$ clipping values, fixed parameters $\xi_1, \xi_2 = [100, 100]$ for the reward definition.
All algorithm candidates use the prior designed for the main study (Table~\ref{tab:finalized_prior}).  
%\sam{About Table 5: I thought the averages would be average brushing quality not average rewards.  Am I missing something?  Before we said we compared RL alg's based on brushing quality not the surrogate reward.   Better to use language proximal outcome or brushing quality than term reward--creates confusion...  Please make two tables so that  each entire table fits on the page...}
%\alt{Hello Susan, we are using average brushing quality and not the average algorithm rewards. I will change the tables and figures to make this more clear. Throughout, I will use the alanauge proximal outcome instead of reward.}

\newgeometry{margin=1cm} % modify this if you need even more space
\begin{landscape}

\begin{table}[]
    \centering
\begin{tabular}{lllllll}
\multicolumn{1}{c}{} & \multicolumn{5}{c}{Smaller Effect Size (Scaling Value $\frac{1}{8}$)} \\
\toprule
\multicolumn{1}{c}{} & \multicolumn{5}{c}{Average Proximal Outcomes} \\
\toprule
ALG\_CANDS &     STAT\_LOW\_R &     STAT\_MED\_R &    STAT\_HIGH\_R & NON\_STAT\_LOW\_R & NON\_STAT\_MED\_R & NON\_STAT\_HIGH\_R \\
\midrule
B = 0.515, Weekly, Full Pooling &     65.365 (0.489) &     66.003 (0.483) &      66.489 (0.483) &         64.172 (0.442) &         64.824 (0.455) &          65.287 (0.453) \\
B = 0.515, Weekly, No Pooling &     64.508 (0.461) &     65.314 (0.468) &      66.077 (0.465) &         63.438 (0.449) &         64.123 (0.449) &          64.875 (0.462) \\
B = 0.515, Daily, Full Pooling &     65.562 (0.482) &     65.929 (0.466) &      66.375 (0.492) &         64.193 (0.447) &         64.966 (0.461) &          65.324 (0.479) \\
B = 0.515, Daily, No Pooling &     64.736 (0.469) &     65.457 (0.467) &      66.129 (0.468) &         63.565 (0.441) &         64.359 (0.443) &          64.900 (0.446) \\
B = 5.15, Weekly, Full Pooling &     65.525 (0.467) &     66.101 (0.487) &      66.617 (0.471) &         64.321 (0.454) &         64.930 (0.446) &          65.516 (0.458) \\
B = 5.15, Weekly, No Pooling &     64.715 (0.473) &     65.582 (0.476) &      66.333 (0.473) &         63.515 (0.436) &         64.324 (0.455) &          65.101 (0.473) \\
B = 5.15, Daily, Full Pooling &     \textbf{65.563 (0.469)} &     \textbf{66.311 (0.473)} &      \textbf{66.733 (0.481)} &         \textbf{64.378 (0.438)} &         \textbf{65.016 (0.459)} &          \textbf{65.575 (0.467)} \\
 B = 5.15, Daily, No Pooling &     64.909 (0.468) &     65.738 (0.482) &      66.184 (0.473) &         63.680 (0.437) &         64.399 (0.448) &          65.091 (0.453) \\
\midrule
\multicolumn{1}{c}{} & \multicolumn{5}{c}{25th Percentile Proximal Outcomes} \\
\midrule
ALG\_CANDS &     STAT\_LOW\_R &     STAT\_MED\_R &    STAT\_HIGH\_R & NON\_STAT\_LOW\_R & NON\_STAT\_MED\_R & NON\_STAT\_HIGH\_R \\
\midrule
B = 0.515, Weekly, Full Pooling &     25.517 (1.157) &     25.446 (1.130) &      \textbf{25.994 (1.186)} &         28.693 (0.957) &         29.532 (1.024) &          28.313 (0.973) \\
B = 0.515, Weekly, No Pooling &     23.841 (1.042) &     24.662 (1.113) &      23.835 (1.088) &         27.618 (0.973) &         28.028 (0.946) &          28.665 (1.014) \\
B = 0.515, Daily, Full Pooling &     \textbf{25.983 (1.109)} &     25.812 (1.163) &      25.183 (1.141) &         28.835 (0.980) &         29.420 (1.021) &          28.822 (1.021) \\
B = 0.515, Daily, No Pooling &     24.230 (1.086) &     24.329 (1.106) &      24.870 (1.089) &         28.565 (0.964) &         28.596 (0.943) &          27.933 (0.979) \\
B = 5.15, Weekly, Full Pooling &     25.420 (1.138) &     25.905 (1.195) &      25.599 (1.123) &         29.132 (0.971) &         29.128 (0.966) &          \textbf{29.248 (0.974)} \\
B = 5.15, Weekly, No Pooling &     24.413 (1.147) &     24.627 (1.121) &      24.496 (1.119) &         28.409 (0.977) &         28.800 (1.001) &          29.027 (0.996) \\
B = 5.15, Daily, Full Pooling &     25.749 (1.173) &     \textbf{25.866 (1.155)} &      25.831 (1.200) &         \textbf{29.541 (0.992)} &         \textbf{29.657 (0.978)} &          28.564 (0.995) \\
B = 5.15, Daily, No Pooling &     24.498 (1.172) &     25.104 (1.227) &      24.693 (1.127) &         27.504 (0.938) &         28.512 (0.979) &          29.143 (0.993) \\
\bottomrule

\multicolumn{1}{c}{} & \multicolumn{5}{c}{Small Effect Size (Scaling Value $\frac{1}{4}$)} \\
\toprule
\multicolumn{1}{c}{} & \multicolumn{5}{c}{Average Proximal Outcomes} \\
\toprule
ALG\_CANDS &     STAT\_LOW\_R &     STAT\_MED\_R &    STAT\_HIGH\_R & NON\_STAT\_LOW\_R & NON\_STAT\_MED\_R & NON\_STAT\_HIGH\_R \\
\midrule
B = 0.515, Weekly, Full Pooling &   71.840 (0.478) &   \textbf{73.249 (0.486)} &    74.433 (0.504) &       69.099 (0.439) &       70.445 (0.444) &        71.385 (0.474) \\
B = 0.515, Weekly, No Pooling &   70.599 (0.466) &   72.011 (0.485) &    73.487 (0.491) &       68.038 (0.438) &       69.473 (0.448) &        70.660 (0.463) \\
B = 0.515, Daily, Full Pooling &   71.918 (0.501) &   73.228 (0.479) &    74.266 (0.491) &      \textbf{ 69.307 (0.449)} &       70.519 (0.461) &        71.502 (0.463) \\
B = 0.515, Daily, No Pooling &   70.826 (0.481) &   72.295 (0.483) &    73.653 (0.495) &       68.387 (0.447) &       69.508 (0.448) &        70.603 (0.458) \\
B = 5.15, Weekly, Full Pooling &   71.946 (0.486) &   73.157 (0.480) &    \textbf{74.532 (0.490)} &       69.297 (0.447) &       70.463 (0.467) &        \textbf{71.643 (0.467)} \\
B = 5.15, Weekly, No Pooling &   70.702 (0.486) &   72.120 (0.494) &    73.722 (0.488) &       68.063 (0.441) &       69.810 (0.458) &        70.768 (0.470) \\
B = 5.15, Daily, Full Pooling &  \textbf{72.088 (0.489)} &   73.176 (0.493) &    74.407 (0.501) &       69.225 (0.451) &       \textbf{70.530 (0.451)} &        71.641 (0.470) \\
 B = 5.15, Daily, No Pooling &   70.826 (0.473) &   72.562 (0.487) &    73.938 (0.501) &       68.352 (0.449) &       69.942 (0.457) &        70.781 (0.469) \\
\midrule
\multicolumn{1}{c}{} & \multicolumn{5}{c}{25th Percentile Proximal Outcomes} \\
\midrule
ALG\_CANDS &     STAT\_LOW\_R &     STAT\_MED\_R &    STAT\_HIGH\_R & NON\_STAT\_LOW\_R & NON\_STAT\_MED\_R & NON\_STAT\_HIGH\_R \\
\midrule
B = 0.515, Weekly, Full Pooling &   33.630 (1.313) &   33.902 (1.280) &    34.660 (1.288) &       \textbf{36.671 (0.988)} &       \textbf{37.347 (1.003)} &        37.477 (1.076) \\
B = 0.515, Weekly, No Pooling &   31.796 (1.215) &   32.979 (1.265) &    33.464 (1.325) &       34.741 (0.980) &       35.565 (1.027) &        36.322 (1.067) \\
B = 0.515, Daily, Full Pooling &   33.744 (1.359) &   34.489 (1.293) &    \textbf{35.404 (1.379)} &       36.614 (0.986) &       36.742 (1.034) &        37.673 (1.056) \\
B = 0.515, Daily, No Pooling &   32.153 (1.187) &   32.767 (1.308) &    33.268 (1.299) &       35.264 (1.022) &       35.372 (1.048) &        35.473 (1.052) \\
B = 5.15, Weekly, Full Pooling &   33.593 (1.242) &   34.182 (1.296) &    34.869 (1.302) &       36.003 (1.050) &       35.985 (1.023) &        37.667 (1.057) \\
B = 5.15, Weekly, No Pooling &   32.328 (1.224) &   32.553 (1.342) &    33.089 (1.314) &       34.422 (1.047) &       36.442 (0.992) &        36.495 (1.031) \\
B = 5.15, Daily, Full Pooling &   \textbf{34.549 (1.245)} &   \textbf{34.914 (1.322)} &    34.655 (1.321) &       36.344 (1.029) &       36.916 (1.004) &        \textbf{37.968 (1.010)} \\
 B = 5.15, Daily, No Pooling &   31.958 (1.182) &   33.238 (1.348) &    34.100 (1.310) &       34.850 (1.032) &       35.743 (1.043) &        36.478 (1.068) \\
\bottomrule
\end{tabular}
\caption{Experiment Results Across 12 Simulation Environment Variants. Value in each parenthesis is the standard error of the mean across 100 simulated trials.}
\label{tab:v2_exp_results}
\end{table}
\end{landscape}
\restoregeometry
%%%% v2 %%%%
\newgeometry{margin=1cm} % modify this if you need even more space
\begin{landscape}
\begin{figure}[ht]
    \centering
    %%%%%%%%% STAT %%%%%%%%%%
\subfloat[Stationary Env. Small Effect Size]{
  \label{fig:stat_small}
    {\includegraphics[width=0.21\textwidth,clip]{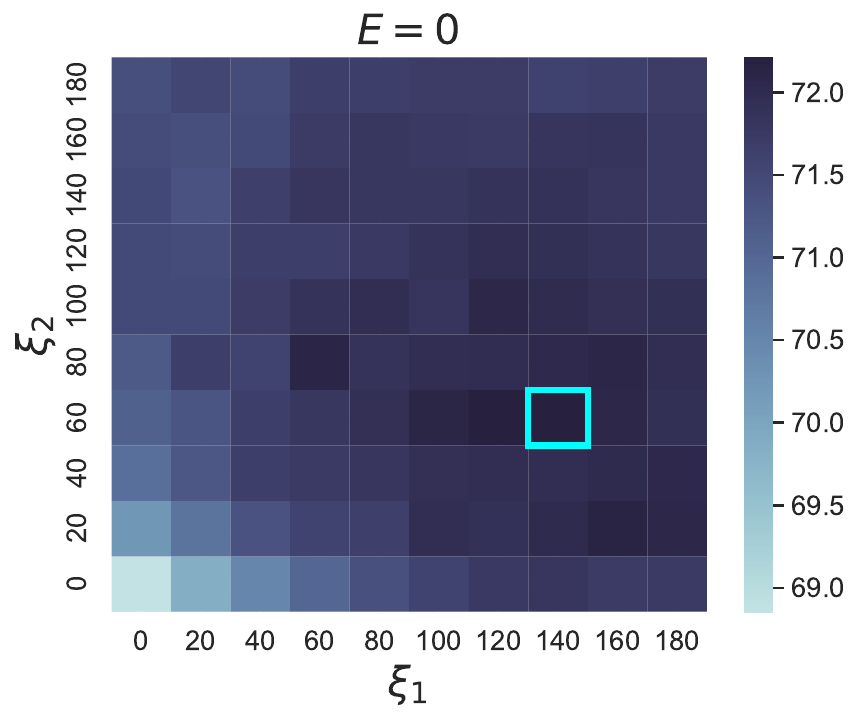}}
    {\includegraphics[width=0.21\textwidth,clip]{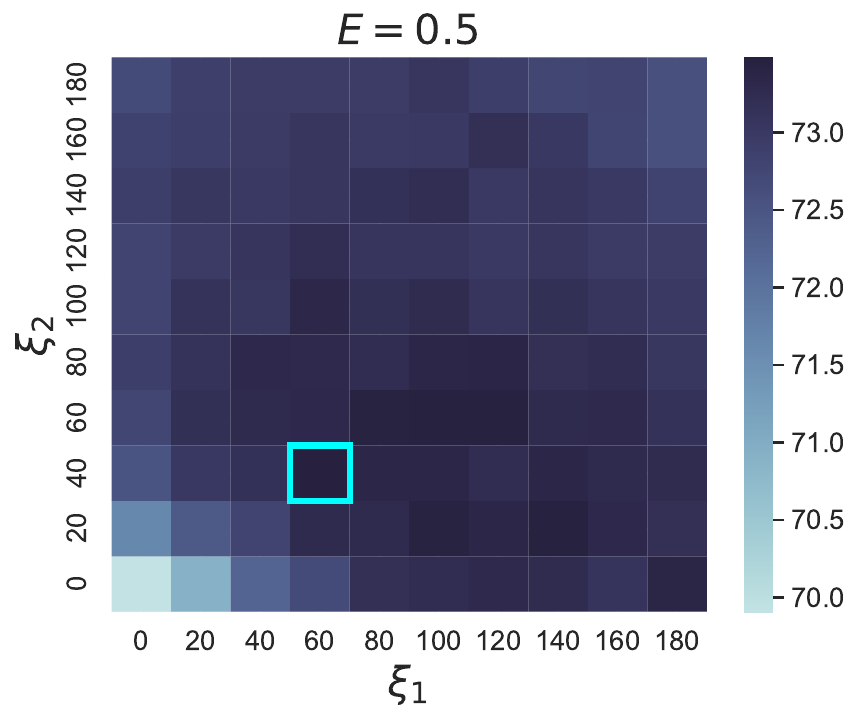}}
    {\includegraphics[width=0.21\textwidth,clip]{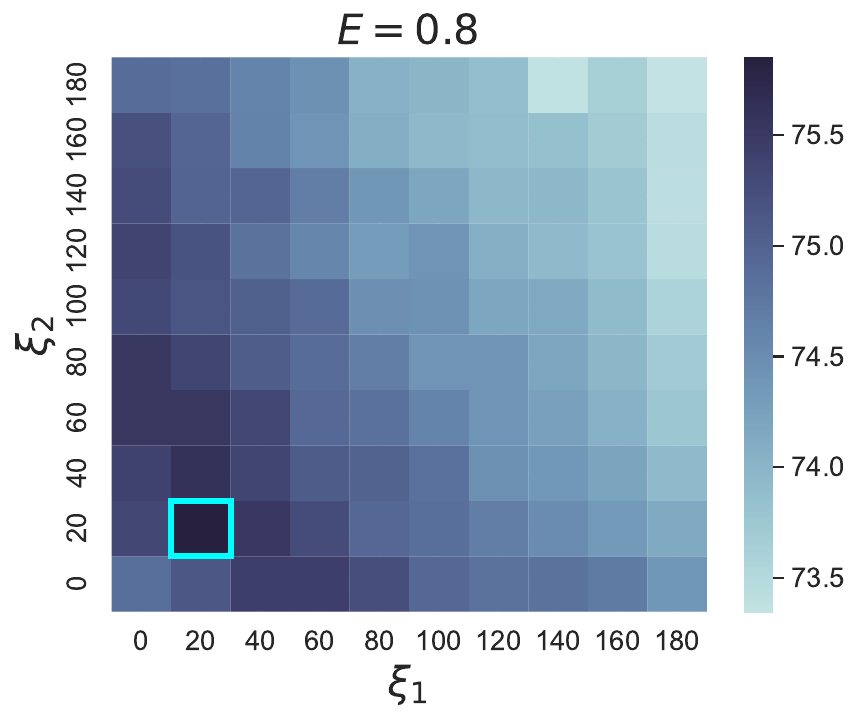}}
    }
    \subfloat[Stationary Env. Smaller Effect Size]{
    {\includegraphics[width=0.21\textwidth,clip]{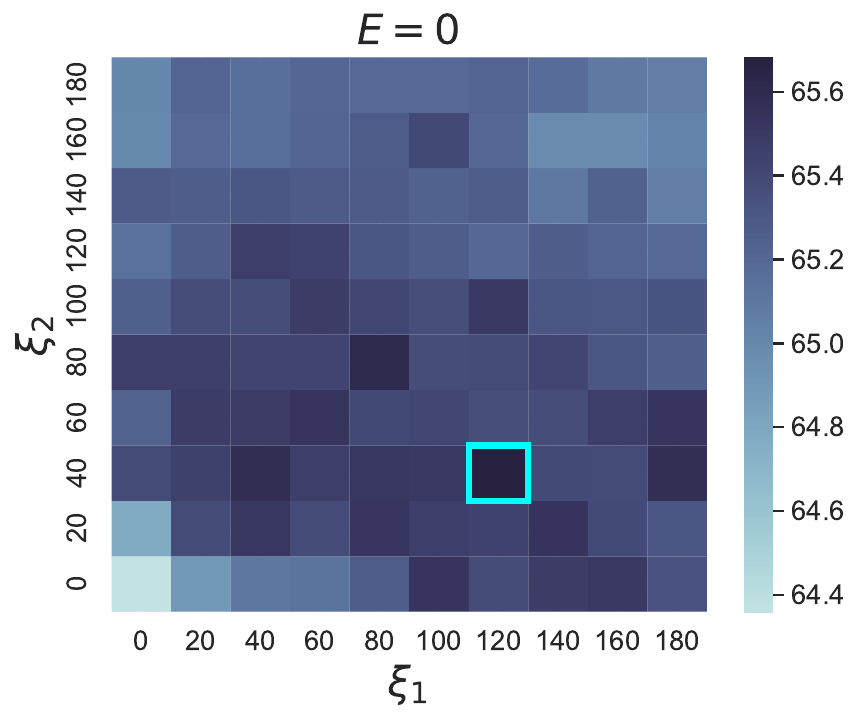}}
    {\includegraphics[width=0.21\textwidth,clip]{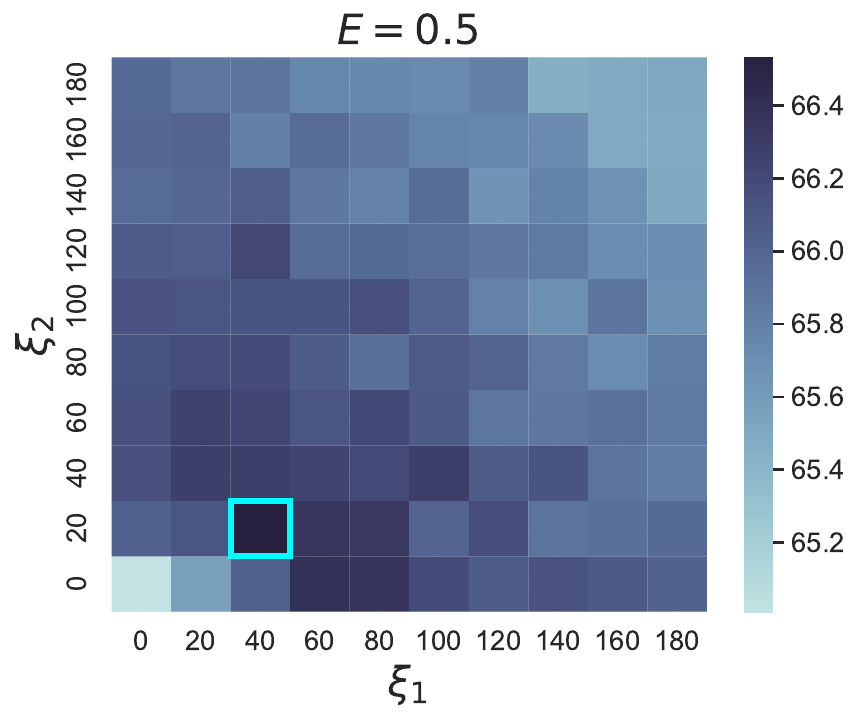}}
    {\includegraphics[width=0.21\textwidth,clip]{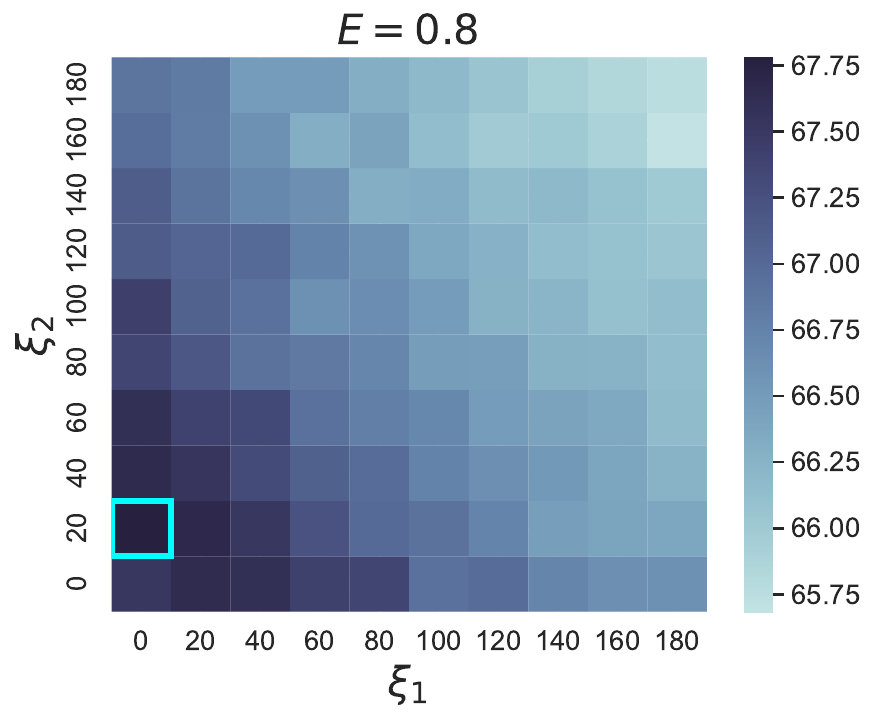}}
    }
    \newline
    %%%%%%%%% NON STAT %%%%%%%%%%
    \subfloat[Non-Stationary Env. Small Effect Size]{
    {\includegraphics[width=0.21\textwidth,clip]{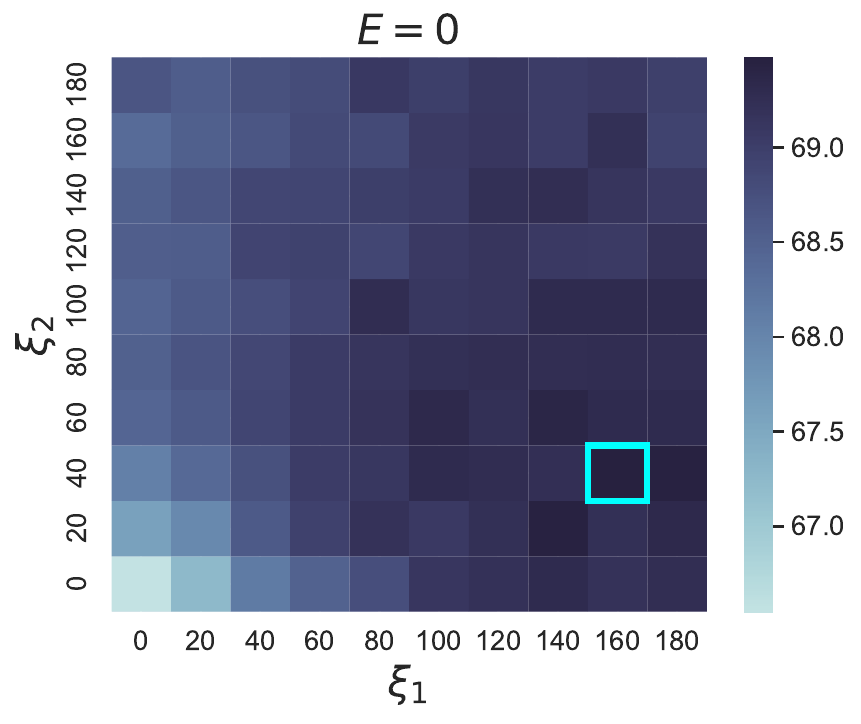}}
    {\includegraphics[width=0.21\textwidth,clip]{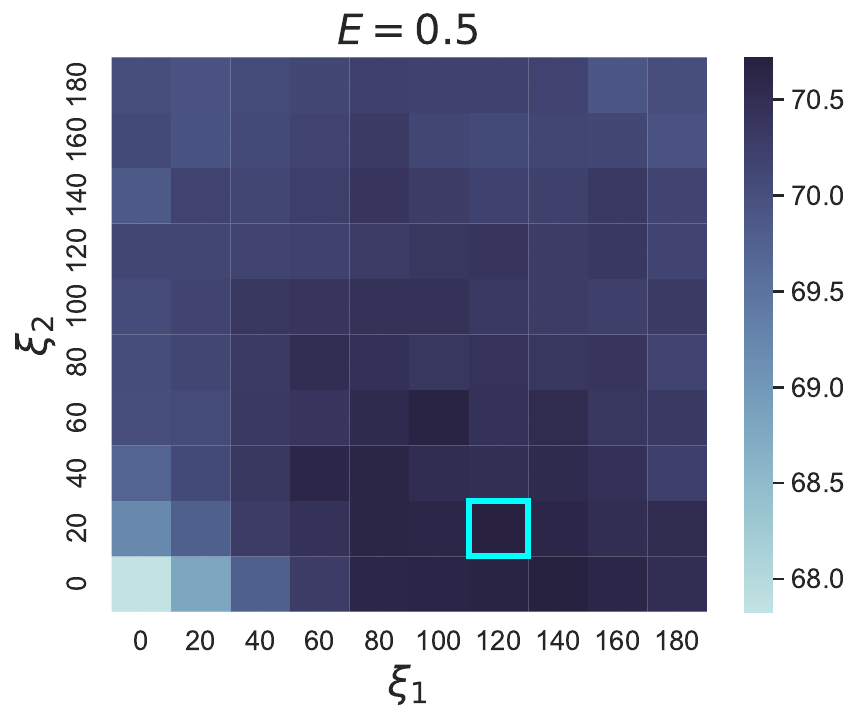}}
    {\includegraphics[width=0.21\textwidth,clip]{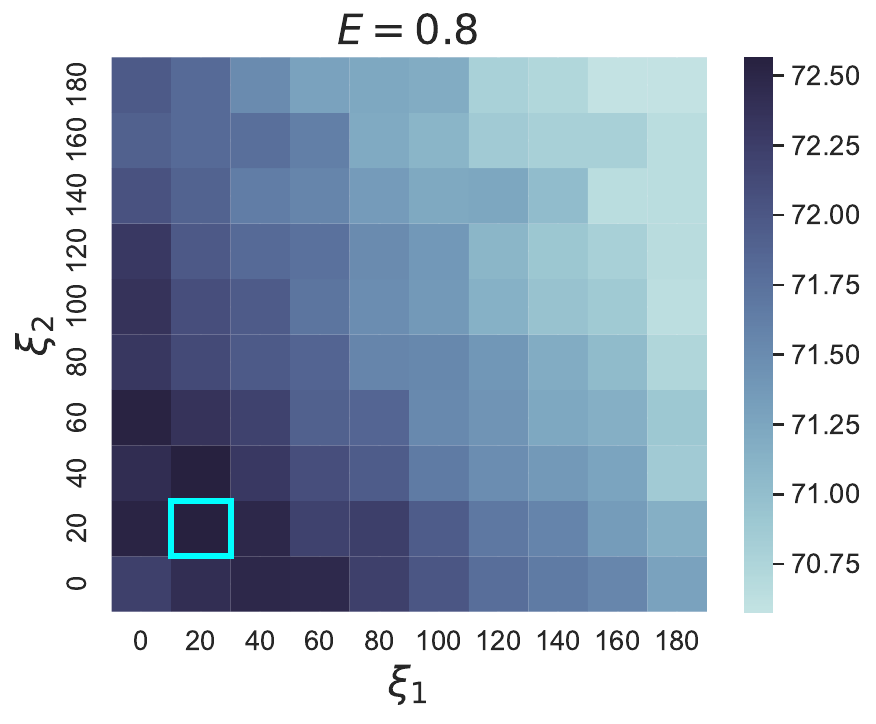}}
    }
    \subfloat[Non-Stationary Env. Smaller Effect Size]{
    {\includegraphics[width=0.21\textwidth,clip]{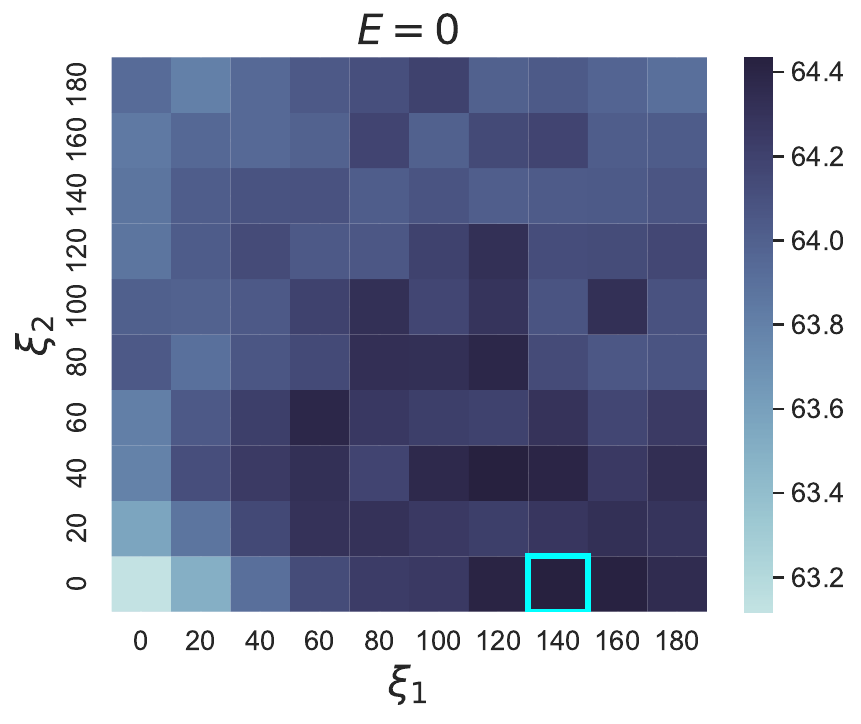}}
    {\includegraphics[width=0.21\textwidth,clip]{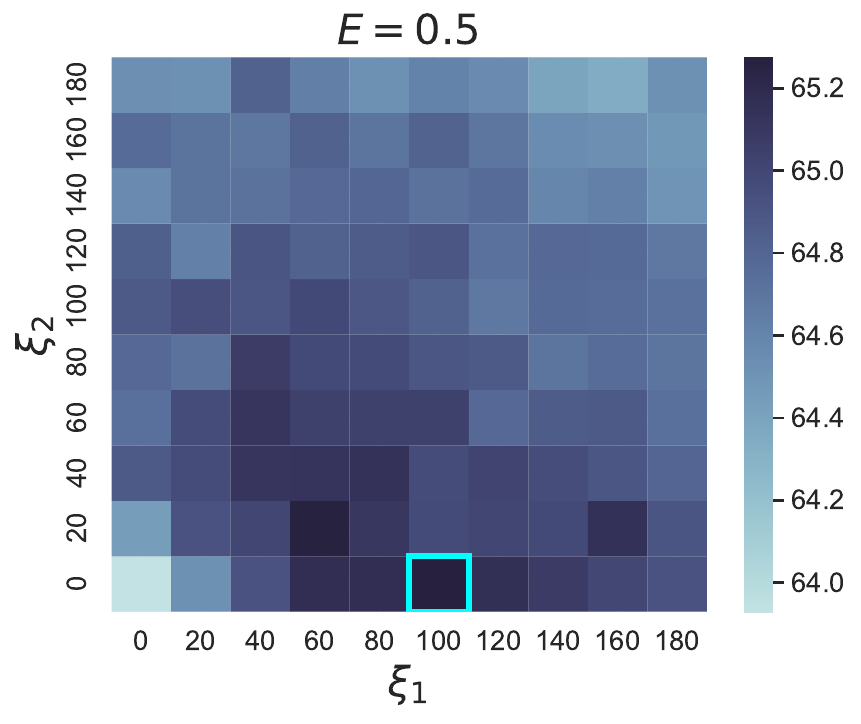}}
    {\includegraphics[width=0.21\textwidth,clip]{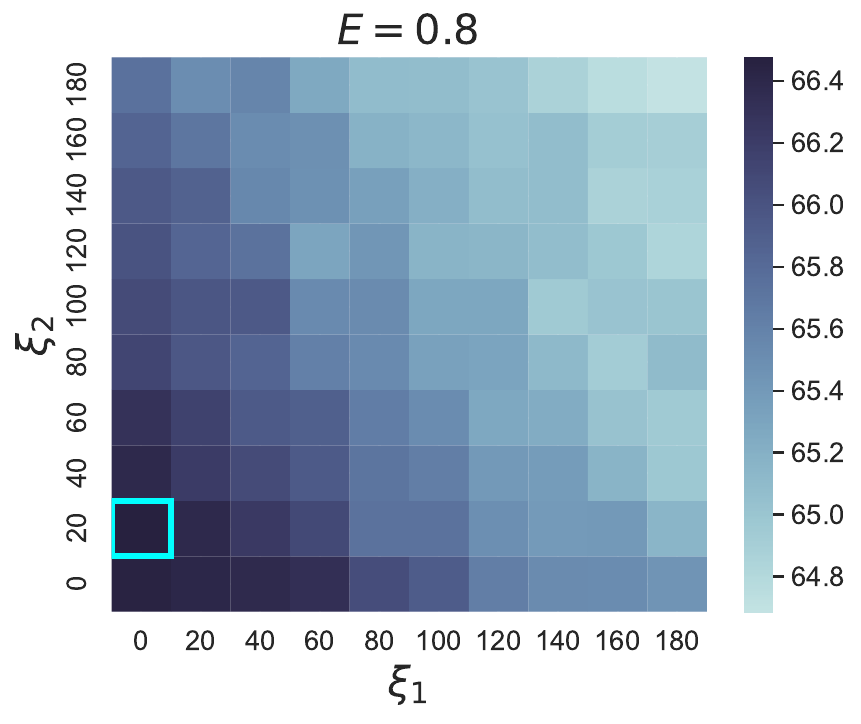}}
    }
    %%%%%%% 25th Percentile %%%%%%%
    %%%%%%%%% STAT %%%%%%%%%%
    \newline
\subfloat[Stationary Env. Small Effect Size]{
  \label{fig:stat_small}
    {\includegraphics[width=0.21\textwidth,clip]{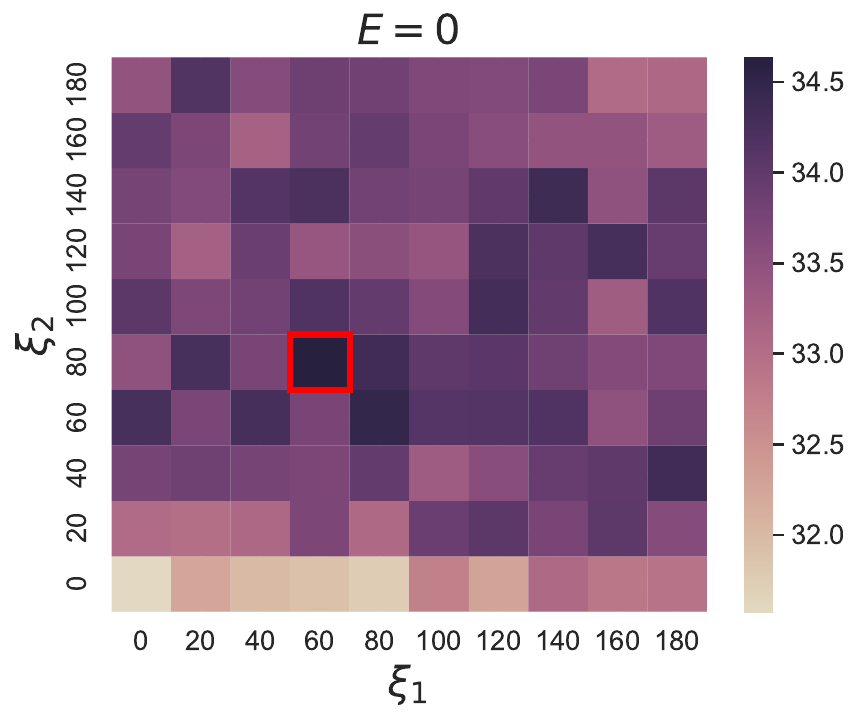}}
    {\includegraphics[width=0.21\textwidth,clip]{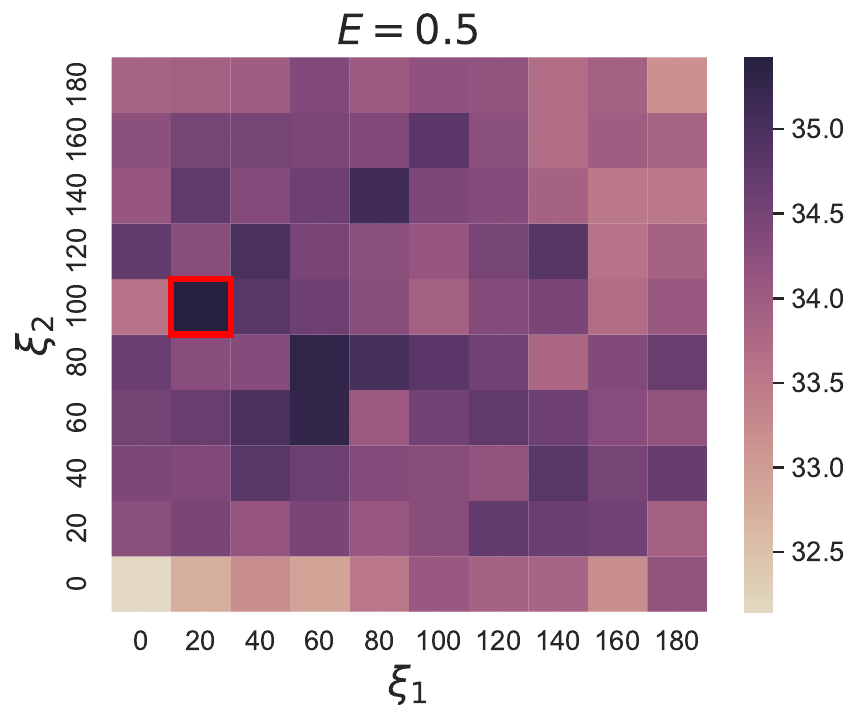}}
    {\includegraphics[width=0.21\textwidth,clip]{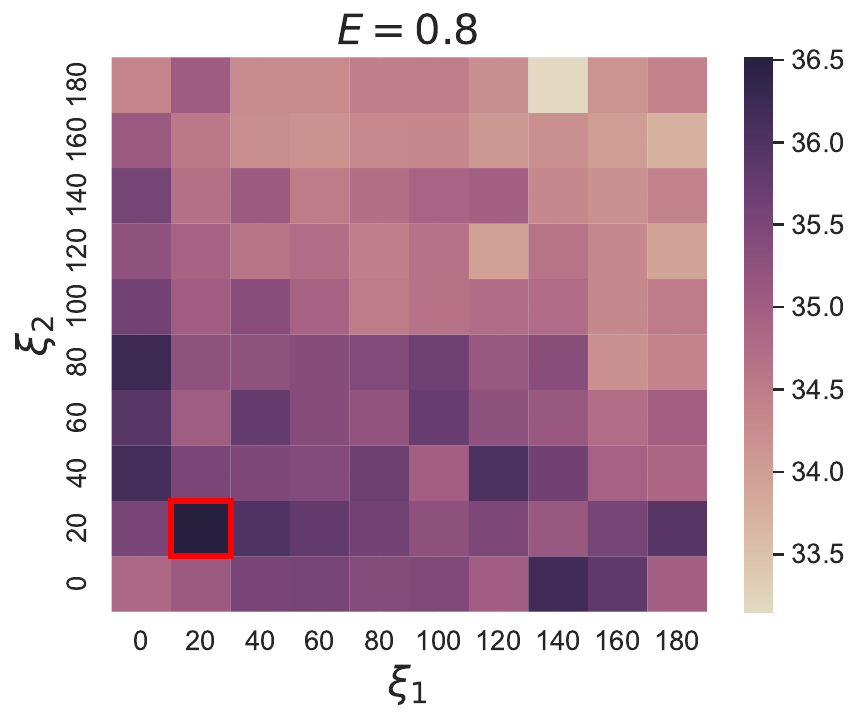}}
    }
    \subfloat[Stationary Env. Smaller Effect Size]{
    {\includegraphics[width=0.21\textwidth,clip]{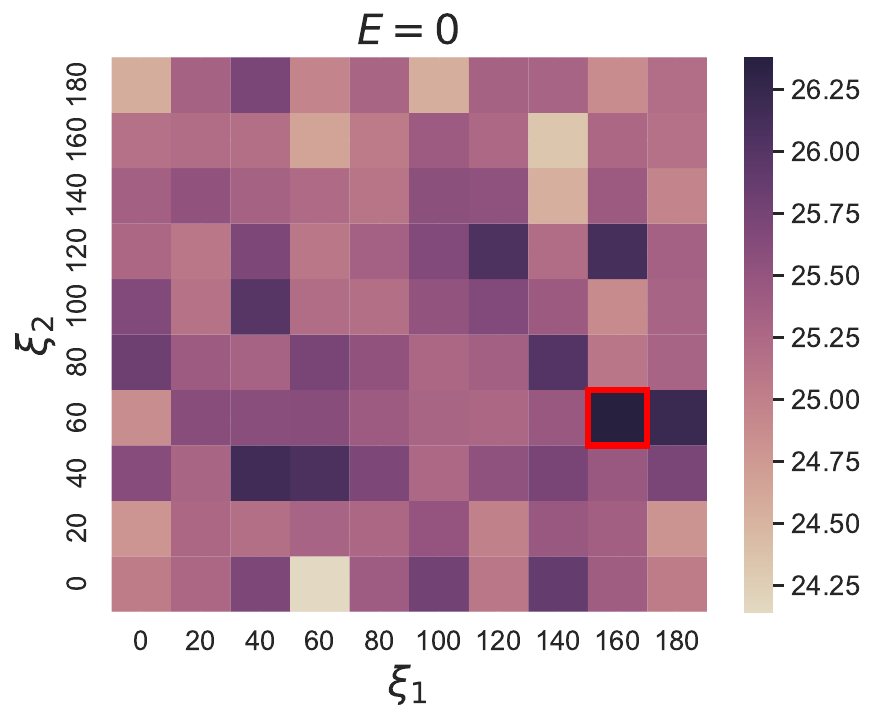}}
    {\includegraphics[width=0.21\textwidth,clip]{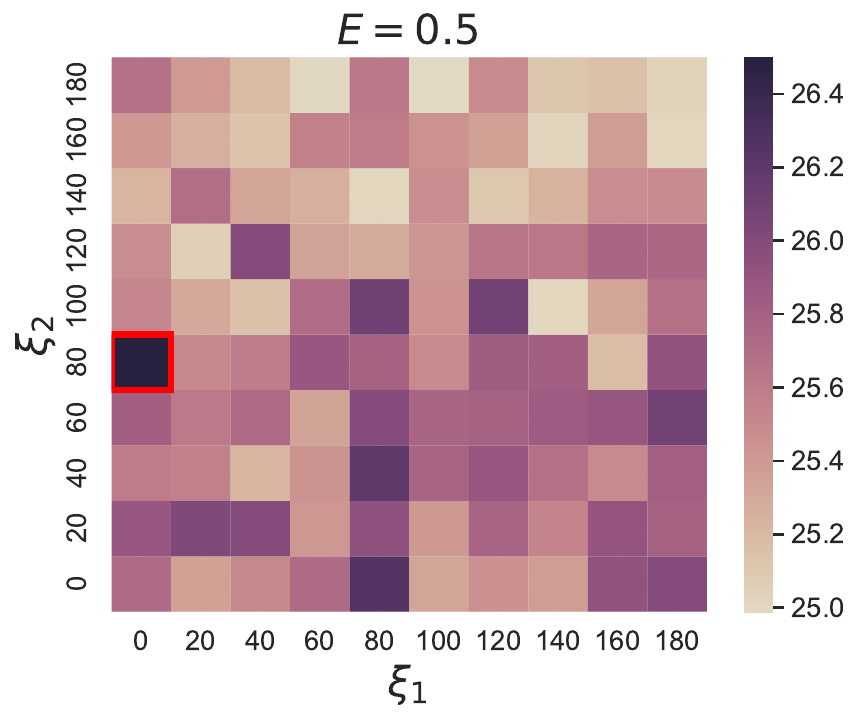}}
    {\includegraphics[width=0.21\textwidth,clip]{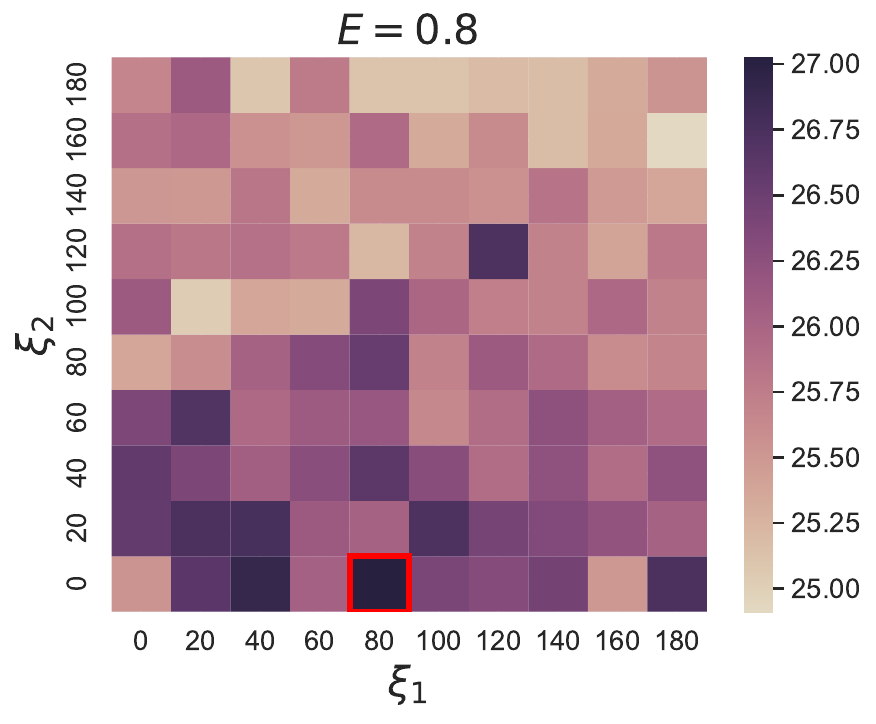}}
    }
    \newline
    %%%%%%%%% NON STAT %%%%%%%%%%
    \subfloat[Non-Stationary Env. Small Effect Size]{
    {\includegraphics[width=0.21\textwidth,clip]{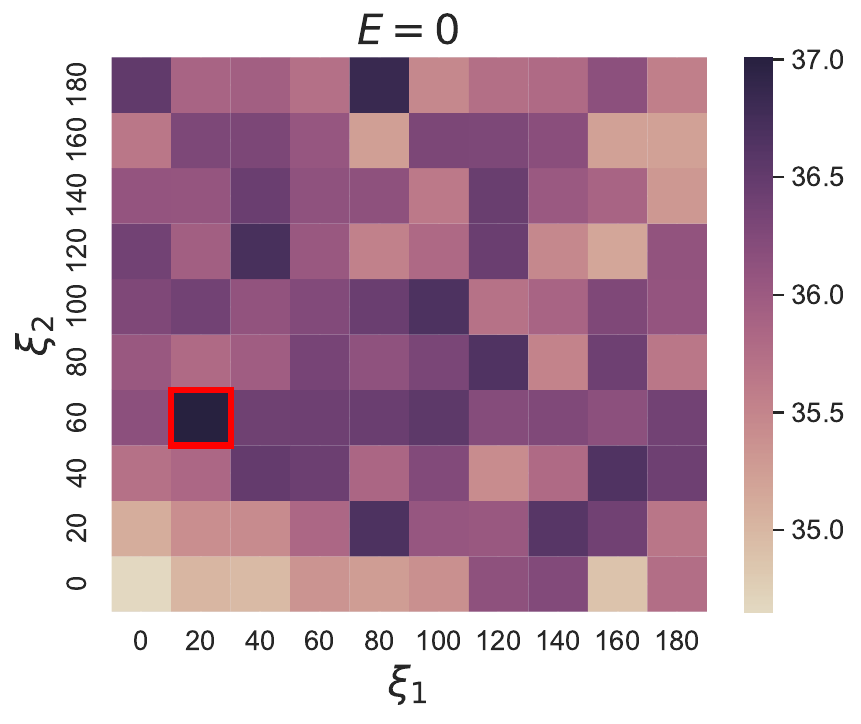}}
    {\includegraphics[width=0.21\textwidth,clip]{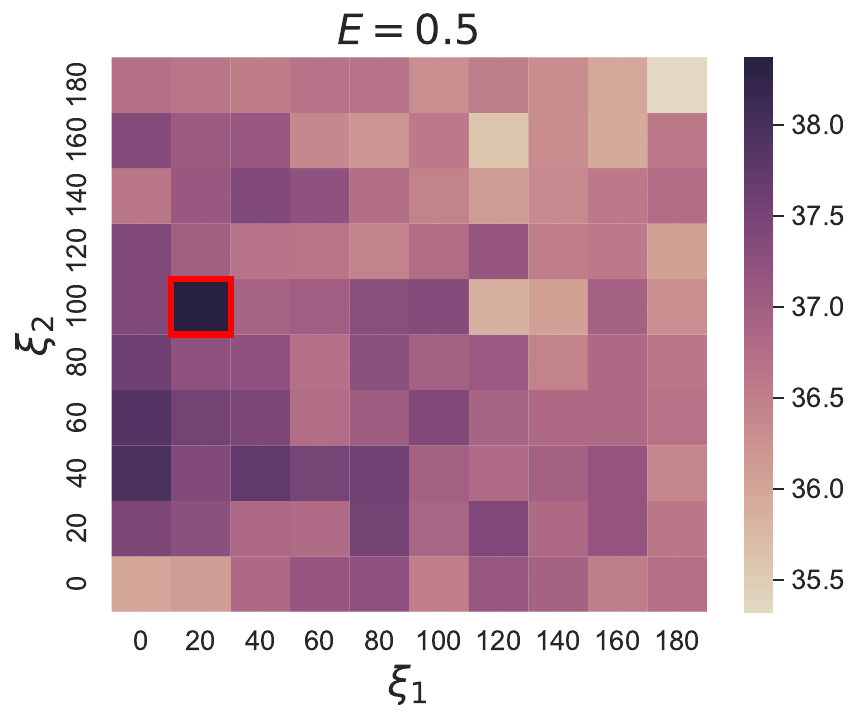}}
    {\includegraphics[width=0.21\textwidth,clip]{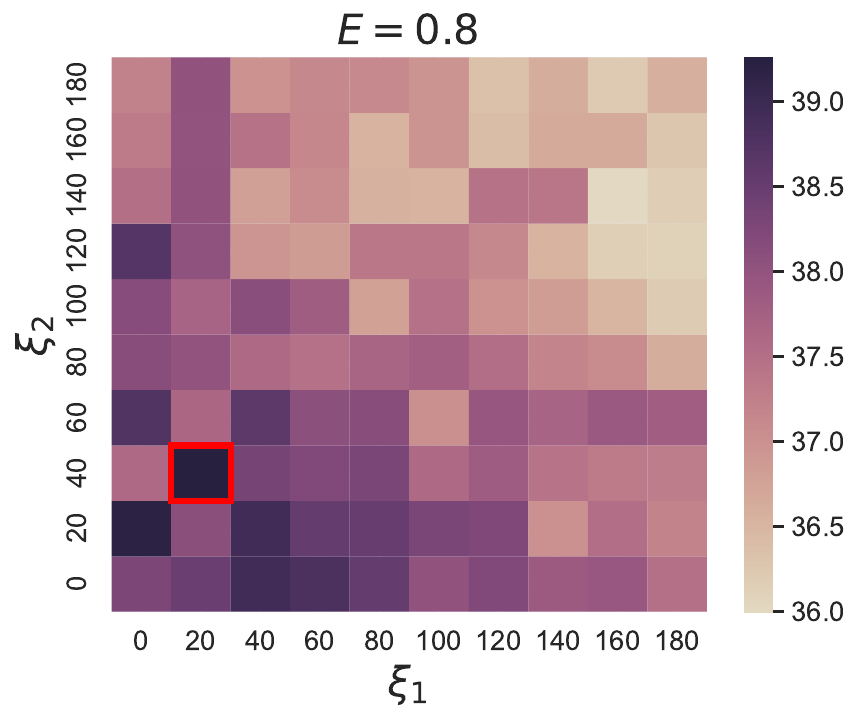}}
    }
    \subfloat[Non-Stationary Env. Smaller Effect Size]{
    {\includegraphics[width=0.21\textwidth,clip]{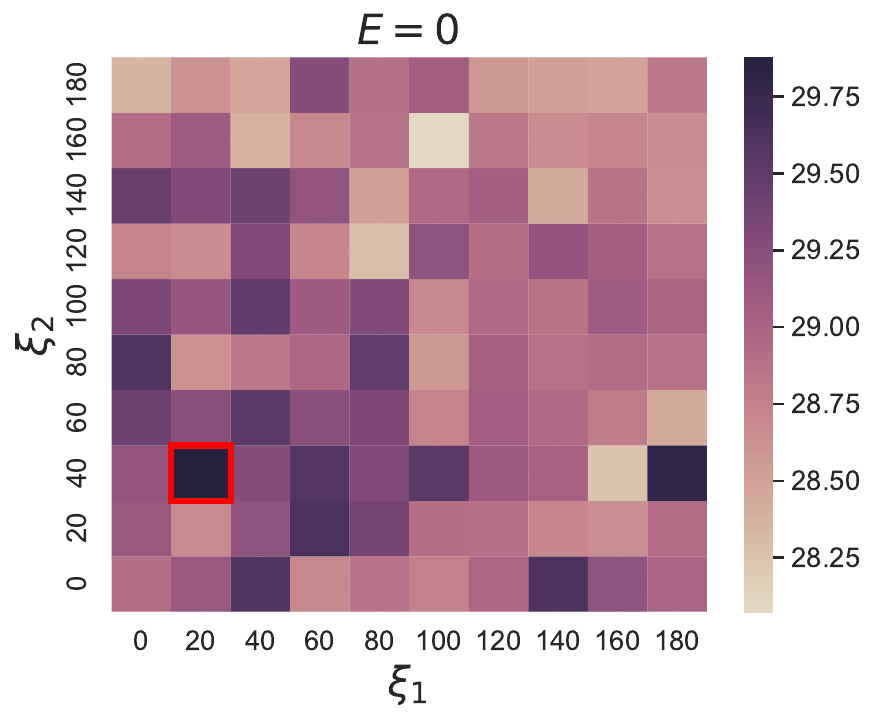}}
    {\includegraphics[width=0.21\textwidth,clip]{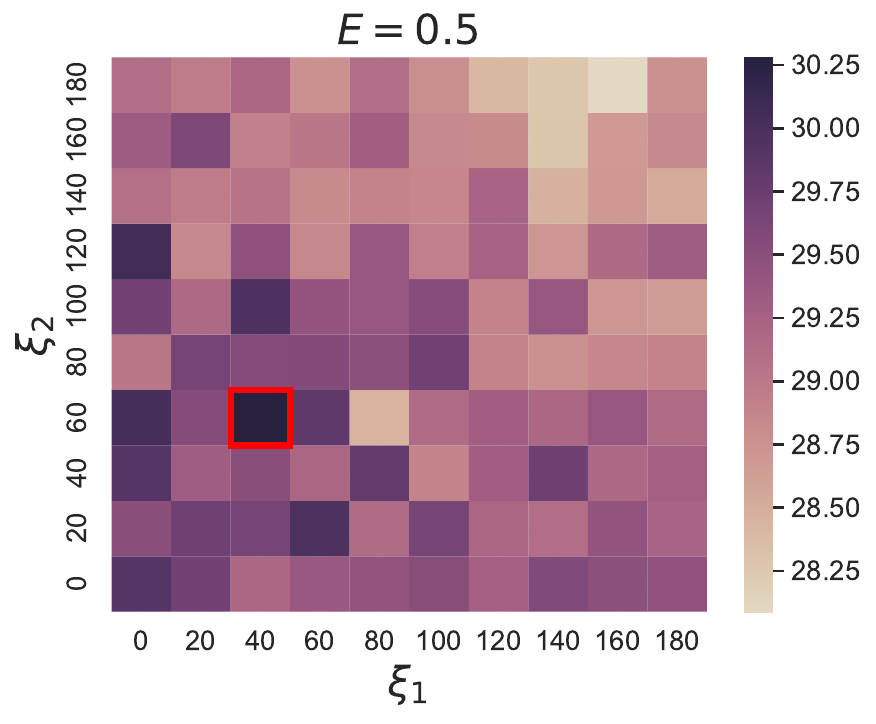}}
    {\includegraphics[width=0.21\textwidth,clip]{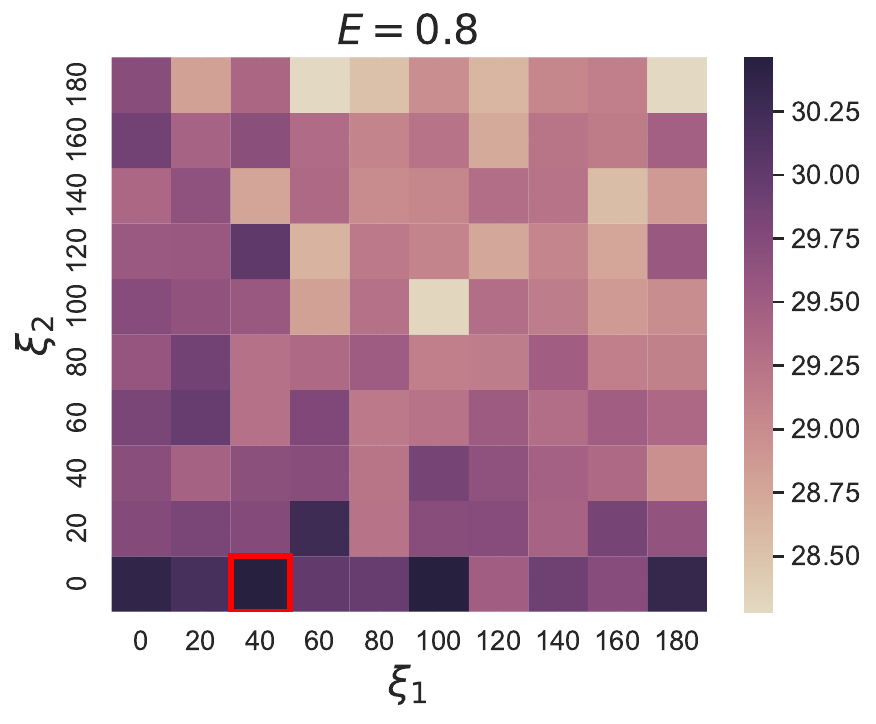}}
    }
    \caption{\textbf{V2 Heatmap of Candidate Values for $\xi_1, \xi_2$.} 
    We evaluate candidate values $\xi_1$, the cost of sending engagement prompts for a high-performing brusher, and $\xi_2$, the cost of sending an engagement prompt regardless of participant performance (See Equation~\ref{cost_term}). We consider two metrics across twelve simulation environment variants (stationary vs. non-stationary base model environment, effect size scales (small and smaller), and effect size shrinkage $E = [0, 0.5, 0.8]$ (small values of $E$ represent low participant robustness to habituation where $E=0$ represents the most severe susceptibility to habituation). The blue grids show simulations evaluated using average $\sum_{t=1}^{T} Q_{i, t}$ across participants and the purple grids show simulations evaluated using $25$th-percentile of $\sum_{t=1}^{T} Q_{i, t}$ across participants. The grid with the highest criteria value is boxed for readability.}
    \label{figs/v2_main_studyheatmaps}
\end{figure}

\end{landscape}
\restoregeometry

\subsection{Hyperparameter Tuning Procedure}
\label{hyper_tuning_proc}
Recall we must select hyperparameters $\xi_1, \xi_2$ in Equation~\eqref{cost_term}. After running experiments (Section~\ref{experiments}) to finalize the design decision on pooling cluster size, update cadence, and slope value of the smoothing allocation function, we keep those decisions fixed and now tune $\xi_1, \xi_2$. We evaluate different values of $\xi_1, \xi_2$ on their ability to maximize the proximal outcome (Section~\ref{brush_quality_def}), $\sum_{t=1}^T Q_{i, t}$, across the twelve simulation environment variants (Section~\ref{sim_env_vars}). In each environment variant, we perform a grid search over the range of possible values of $\xi_1, \xi_2$. 
%For each algorithm variant (value of $\xi_1, \xi_2$) and simulation environment pair, 
We consider the following evaluation metrics:

\begin{enumerate}
    \item Average cumulative OSCB across all participants, $\frac{1}{N} \sum_{i=1}^N \sum_{t=1}^T Q_{i, t}$
    \item 25th-percentile of cumulative OSCB, $\sum_{t=1}^T Q_{i, t}$, across all participants
\end{enumerate}

For each environment and $\xi_1, \xi_2$ value pairing, we again simulate a study with $N=70$ participants over $T=140$ decision points as described in Section~\ref{experiments}. For each environment variant, we generate two grids corresponding to the two evaluation criteria described above. Each square in a grid represents a criterion evaluated on a simulated study using values $\xi_1, \xi_2$ for the cost term, averaged across $100$ Monte Carlo simulated trials. Figure \ref{figs/v2_main_studyheatmaps} shows heat maps of the 2 evaluation criterion values for different values of $\xi_1, \xi_2$ across 12 environment variants.
\subsubsection{Final Algorithm Decisions}
\label{final_alg_decisions}
Using simulations (Section~\ref{experiments}) we made the following decisions. The first three decisions were informed by the results in Table~\ref{tab:v2_exp_results} and the final hyperparameterer values were determined using the results in Table~\ref{figs/v2_main_studyheatmaps}.
\begin{itemize}
    \item Pooling Cluster Size: Full Pooling (Reason: In our experiments, full pooling yielded higher average and 25th percentile of average reward across all environment variants.)
    \item Update Cadence: Weekly (Reason: In our experiments, there very little difference in average OSCB between a daily and update cadence across all environment variants. The team decided on a weekly cadence to simplify the  after-study analyses / computational reasons. Because of the similar performance, we made decisions using additional considerations.)
    \item Parameters of Smoothing Allocation Function
    \begin{itemize}
        \item Slope Value $B = 0.515$ (Reason: Even though steeper seems to do a bit better overall, it's not significantly different from the less steep. We chose the less steep slope to increase power in the after-study analyses.)
    \end{itemize}
    \item 
    %\sam{SUSAN CHECK: could you please check my english for the hyperparameter tuning choices?}
    Hyperparameters for Reward Definition: $\xi_1, \xi_2 = [80, 40]$ (Reason: The initial experiments in \cite{trella2023reward} were based off of ROBAS 3, which was a study where each participant was in the study for more than 70 days. Results indicated the need to penalize based on prior dosage resulting in the initial $\xi_1, \xi_2 = [100, 100]$ values. However analyses of  the Oralytics pilot study data  show a positive correlation between prior dosage and current brushing duration indicating no need to penalize based on prior dosage (Figure~\ref{fig:ac_model_standard_eff}). But the pilot had only 9 people and also provided a nonsensical negative effect of sending an engagement prompt, thus reducing confidence in the positive correlation. Thus we reduced both tuning parameters to values that still achieved high average reward (Figure~\ref{figs/v2_main_studyheatmaps}).)
\end{itemize}

% \paragraph{Observations on Simulation Results}
% \begin{itemize}
%     \item \bo{Pooling:} I looks like overall that full pooling is better than no pooling. The exception is for V3 25th percentile simulations; the differences between algorithms is very small though.
%     \item \bo{Daily vs. Weekly Updates:} Daily looks like it in general does a little bit better in V2 simulations, but weekly might be a bit better in V3 setting.
%     The team decided on a weekly cadence to ensure stability for after-study analyses / computational reasons. Because of the similar performance we made decisions using additional considerations. \alt{ANNA TODO: change!! first Sunday after  the 15th participant entered.}
%     \item \bo{Steepness of Slope:} Even though steeper seems to do a bit better overall, it's not significantly different from the less steep. We chose the less steep slope to increase power in the after-study analyses.
% \end{itemize}
\subsection{Varying Prior Sampling Period}
Recall that the algorithm undergoes a prior sampling period where the algorithm initially samples actions from a prior distribution (Section~\ref{onboarding_proc}). The prior sampling period continues until one week after the 15th participant starts the study, where 15 was chosen in discussion with domain experts. With the finalized RL algorithm (Section~\ref{final_alg_decisions}), we ran a final set of simulations comparing the average OSCB achieved when the prior sampling period instead continues until one week after the 5th participant starts the study (i.e., posterior sampling starts earlier in the study). Results are in Table~\ref{tab_prior_sampling_duration}. As part of study activities, we will also conduct additional sensitivity analysis to ensure that the after-study confidence intervals are accurate even if we started the RL period earlier.

\begin{table}[!ht]
    \centering
    \begin{tabular}{ccc}
    & \multicolumn{2}{c}{Smaller Effect Size (Scaling Value $\frac{1}{8}$)}\\
    \toprule
    & \multicolumn{2}{c}{Average Proximal Outcomes}\\
    \hline
      Environments & Longer Period & Shorter Period \\      
        \hline
        STAT\_LOW\_R & 65.513 (0.471) & 65.494 (0.468) \\
        STAT\_MED\_R & 66.189 (0.484) & 66.262 (0.475) \\
        STAT\_HIGH\_R & 66.976 (0.486) & 66.833 (0.471) \\
        NON\_STAT\_LOW\_R & 64.183 (0.452) & 64.259 (0.443) \\
        NON\_STAT\_MED\_R & 65.144 (0.457) & 65.066 (0.450) \\
        NON\_STAT\_MED\_R & 65.709 (0.452) & 65.867 (0.458) \\
        \hline
    & \multicolumn{2}{c}{25th Percentile Proximal Outcomes}\\
    \hline
    % & Longer Period & Shorter Period \\
        STAT\_LOW\_R & 25.691 (1.168) & 25.959 (1.158)  \\
        STAT\_MED\_R & 26.199 (1.202)  & 25.289 (1.140)\\
        STAT\_HIGH\_R & 26.627 (1.177)  & 26.498 (1.184) \\
        NON\_STAT\_LOW\_R & 29.296 (0.989) &  28.291 (0.942)\\
        NON\_STAT\_MED\_R & 29.802 (1.017) & 29.422 (0.978) \\
        NON\_STAT\_MED\_R & 29.236 (1.008) & 29.511 (1.007) \\
        \hline
        & \multicolumn{2}{c}{Small Effect Size (Scaling Value $\frac{1}{4}$)}\\
        \toprule
    & \multicolumn{2}{c}{Average Proximal Outcomes}\\
    \hline
        STAT\_LOW\_R & 71.807 (0.483) & 71.921 (0.491) \\
        STAT\_MED\_R & 73.364 (0.489) & 73.311 (0.467) \\
        STAT\_HIGH\_R & 74.993 (0.496) & 74.867 (0.492)\\
        NON\_STAT\_LOW\_R & 69.115 (0.440) & 69.055 (0.458) \\
        NON\_STAT\_MED\_R & 70.646 (0.463) & 70.604 (0.460) \\
        NON\_STAT\_MED\_R & 71.949 (0.459) & 71.803 (0.464) \\
        \hline
    & \multicolumn{2}{c}{25th Percentile Proximal Outcomes}\\
    \hline
        STAT\_LOW\_R & 33.975 (1.245) & 34.018 (1.308) \\
        STAT\_MED\_R & 34.338 (1.254) & 34.459 (1.312) \\
        STAT\_HIGH\_R & 35.681 (1.402)  & 35.899 (1.303)\\
        NON\_STAT\_LOW\_R & 35.852 (1.058) & 35.732 (1.107) \\
        NON\_STAT\_MED\_R & 37.549 (1.063) &  37.690 (1.038)\\
        NON\_STAT\_MED\_R & 38.290 (1.076) &  37.213 (1.064)\\
    \end{tabular}
    \caption{Experiment results for varying durations of the prior sampling period. ``Longer period" refers to the prior sampling period lasting until one week after the 15th participant enters the study and ``Shorter period" refers to the 5th.}
\label{tab_prior_sampling_duration}
\end{table}

%%%%%%%% SIMULATION ENV. %%%%%%%
\section{Simulation Environments}
\label{app:sim_env}
% \sam{we need to explain for each environment variant, we generate 31 simulated participants, based on each of the 31 participants in ROBAS 3.   I think most of the text below is participant-specific.  So we are building a simulated participant $i$ but this is pretty unclear...  Since we have 12 environment variants, essentially we are building 12 versions of the ith ROBAS 3 participant...Is this correct?}
% \alt{yes Susan you are correct!}
To create $N$ participant environment models we sample, with replacement from 31 simulated participants, based on each of the 31 participants in ROBAS 3. Recall we also consider 12 total environment variants. Therefore we are building 12 versions of each ROBAS 3 participant environment model.

Each participant environment model can further be represented by the following components:

\begin{itemize}
    \item Outcome Generating Process (i.e., brushing quality given state and action)
    \item Responsivity To Actions (i.e., delayed effect of current action on participant responsivity to future actions)
    \item App Engagement Behavior
\end{itemize}

In this section, we detail how we developed the simulation environment used to inform, test, and evaluate the design of the Oralytics RL algorithm. See Table~\ref{tab:env_versions} for a high-level overview of the properties of the simulation environment.

\begin{table}[h]
    \centering
    \begin{tabular}{cc}
     Property &  \\
    \toprule
    Data Set & ROBAS 3 \\
    Simulates App Engagement? & Yes (See Section~\ref{sim_env_variants_open_app}) \\
    Simulates App Opening Issue? & Yes (See Section~\ref{sim_env:app_open_issue}) \\
    Constructing Treatment Effect & Imputation  (See Section~\ref{treatment_eff_sizes})\\
    Environment Feature Space & (See Section~\ref{baseline_features}) \\
    Recruitment Rate & 5 participants per 2 weeks \\
\end{tabular}
\caption{{\textbf{Properties of Simulation Environment}}. ``Data Set" refers to the data set used to fit the environment base model (Section~\ref{base_model_def} details building the environment base model). ``Simulates App Opening Issue'' refers to simulating participants only receiving the most recent action if they opened the app the day before. See Section~\ref{app_open_issue} for further details about the app opening issue.}
\label{tab:env_versions}
\end{table}

\subsection{Outcome Generating Process} Since we built the simulation environment using ROBAS 3 data which had no data under action 1, we decomposed the outcome-generating process for each participant into two components: 1) we fit a baseline outcome (i.e., the OSCB under action $A_{i,t} = 0$) (Section~\ref{base_model_def}) for each participant using ROBAS 3 data and 2) we impute a treatment effect to model OSCB under action $A_{i,t} = 1$ (Section~\ref{treatment_eff_sizes}).

% \subsubsection{Building the Environment Base Model}
% \label{env_base_model}
% The baseline feature space is found in Section~\ref{baseline_features}. %We only use the ROBAS 2 dataset for fitting the prior of the algorithm. 
\subsubsection{Baseline Feature Space}
\label{baseline_features}
We introduce the baseline feature space for the environment base models. These features were selected  using domain expert knowledge from behavioral health and dentistry. 

\begin{enumerate}
    \item Time of Day (Morning/Evening) $\in \{0, 1\}$
    \item Prior Day Total OSCB (Normalized) $\in \mathbb{R}$ 
    \item Weekend Indicator (Weekday/Weekend) $\in \{0, 1\}$
    \item Proportion of Non-zero Brushing Sessions Over Past 7 Days $\in [0, 1]$
    \item Day in Study (Normalized) $\in [-1,1]$
    \item Bias / Intercept Term $\in \mathbb{R}$
\end{enumerate}

We use these features to generate two types of base reward environments (Stationary and Non-Stationary). The Stationary model of the base environment uses the state function $g(S_{i,t}) \in \mathbb{R}^5$ that includes all features above, except for ``Day in Study". The Non-Stationary model of the base environment uses state $g(S_{i,t}) \in \mathbb{R}^6$ that corresponds to all of the above features.

\subsubsection{Normalization of Environment State Features}
\label{normalizations}
We normalize features to ensure that all state features are within a similar range. The Past OSCB feature is normalized using z-score normalization (subtract mean and divide by standard deviation) using ROBAS 3 OSCB data. The Day in Study feature is normalized based on Oralytics' anticipated 70-day study duration (range is still $[-1,1]$). Notice that both fitting simulated participants models (Section~\ref{fitting_env_base_models}) and generating outcomes (Section~\ref{reward_gen_process}) use the same Day in Study normalization. %\sam{reader will not understand prior sentence unless we tell reader where normalization is for generating rewards and where normalization is for fitting a model to each of the 31 robas 3 participants.} %\alt{good catch Susan, I added more description and links above}

$$\text{Normalized Past OSCB in Seconds} = (\text{OSCB} - 154) / 163 $$

$$\text{Normalized Day in Study} = (\text{Day} - 35.5) / 34.5$$

\subsubsection{Environment Base Model Definitions}
\label{base_model_def}
\begin{figure}[!h]
    \centering
\includegraphics[width=0.59\textwidth]{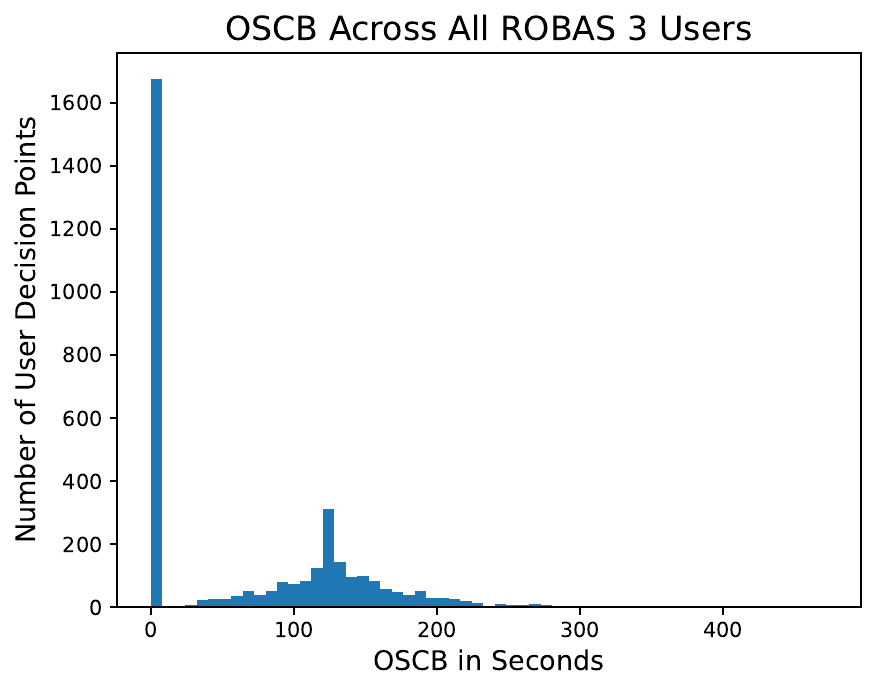}
    \caption{\textbf{Histogram of OSCB in ROBAS 3.} OSCB across all 31 participants in ROBAS 3 across 140 brushing windows (2 brushing windows per day for 70 days). Since the ROBAS 3 study lasted for 90 days, but each participant for Oralytics will only be in the study for 70 days, we only take the first 70 days of data for each participant in ROBAS 3. 
    %Notice that across all participants and brushing windows, a \alt{ANNA TODO} had a OSCB of zero seconds.
    Notice that the ROBAS 3 data set is highly zero-inflated.
%     \sam{Anna, We should try to include two  histograms--otherwise people will wonder why we are fitting 0 inflated models.  Histogram 1:  OSCB over all 31x140 ROBAS 3 participant-decision points?   The y axis would be fraction of participant-decision points and the x axis would be the OSCB in seconds. 
% Histogram 2: pick a representative ROBAS 3 participant and now the y axis is the fraction of 140 decision points and the x axis is the OSCB in seconds.   } \alt{I will work on creating those histograms!} 
    }
    \label{fig:robas3_hist}
\end{figure}

$g(S_{i, t})$ is the baseline feature vector of the current state defined in Section~\ref{baseline_features}. $w_{i,b}, w_{i,p}, w_{i,\mu}$ are participant-specific weight vectors, $\sigma_{i,u}^2$ is the participant-specific variance for the normal component, and $\mathrm{sigmoid}(x) = \frac{1}{1 + e^{-x}}$ is the sigmoid function. To generate OSCB for participant $i$ at decision time $t$, we consider a zero-inflated Poisson model and a hurdle model. We consider both models due to the zero-inflated nature of OSCB found in the ROBAS 3 data set; see Figure~\ref{fig:robas3_hist}. The two models are defined as:

\paragraph{(1) Zero-Inflated Poisson Model for OSCB:}
$$
Z \sim \text{Bernoulli}\left(1 - \mathrm{sigmoid}(g(S_{i, t})^T w_{i,b}) \right)
$$
$$
Y \sim \text{Poisson} \left( \exp \left( g(S_{i, t})^T w_{i,p} \right) \right)
$$
$$
\text{OSCB} : Q_{i, t} = ZY
$$

\paragraph{(2) Hurdle Model with Square Root Transform for OSCB}
$$
Z \sim \text{Bernoulli} \left(1 - \mathrm{sigmoid} \left( g(S)^T w_{i,b} \right) \right)
$$
$$
Y \sim \mathcal{N} \left( g(S)^T w_{i,\mu}, \sigma_{i,u}^2 \right)
$$
$$
\text{OSCB} : Q_{i, t} = ZY^2
$$

\subsubsection{Fitting the Environment Base Model}
\label{fitting_env_base_models}
% \sam{Did we actually use both of these participant-specific models?..   I ask because in design paper we only used the zero-inflated poisson model.} 
% \alt{Hello Susan, yes we used both ZIP and hurdle.}
For each model class that models OSCB under no intervention ($A_{i, t} = 0$), we fit participant-specific models (separately for each of the 31 ROBAS 3 participants). All models were fit using MAP with a prior $w_{i,b}, w_{i,p}, w_{i,\mu} \sim \mathcal{N}(0, I)$ as a form of regularization because we have sparse data for each participant. Weights were chosen by running random restarts and selecting the weights with the highest log posterior density.

\paragraph{Fitting Hurdle Models:}
For fitting hurdle models for participant $i$, we fit the Bernoulli component and the nonzero brushing duration component separately. Set $Z_{i,t} = 1$ if the original observation $Q_{i, t} > 0$ and $0$ otherwise. We fit a model for this Bernoulli component. We then fit a model for the normal component to the square root transform $Y_{i, t} = \sqrt{Q_{i, t}}$ of the $i$th participant's nonzero OSCB. 

\paragraph{Fitting Zero-Inflated Poisson Models:} For the zero-inflated Poisson model, we jointly fit parameters for both the Bernoulli and the Poisson components.

\subsubsection{Selecting the Model Class For Each Participant}
\label{env_model_selection}
To select the model class for participant $i$, we fit both model classes specified above using participant $i$'s data from ROBAS 3 We then chose the model class that had the lowest RMSE. Namely, we choose the model class with the lowest $L_i$, where:

$$
L_i := \sqrt{\sum_{t = 1}^T (Q_{i, t} - \hat{\mathbb{E}}[Q_{i, t} | S_{i, t}])^2}
$$
Recall that $Q_{i,t}$ is the OSCB in seconds for participant $i$ at decision point $t$.
Definitions of $\hat{\mathbb{E}}[Q_{i, t} | S_{i, t}]$ for each model class are specified below in Table~\ref{table/mean_for_model_classes}.

\begin{table}[!h]
    \centering
\begin{tabular}{c|c}
 \hline
 \textbf{Model Class} & $\widehat{\mathbb{E}}[Q_{i, t} | S_{i, t}]$ \\
 \hline
    Zero-Inflated Poisson & $\left[1 - \mathrm{sigmoid} \left(g(S_{i, t})^Tw_{i,b} \right) \right] \cdot \exp \left( g(S_{i, t})^Tw_{i,p} \right)$ \\
    Hurdle (Square Root) & $\left[1 - \mathrm{sigmoid}\left( g(S_{i, t})^T w_{i,b} \right) \right] \cdot \left[\sigma_{i,u}^2 + (g(S_{i, t})^T w_{i,\mu})^2\right]$  \\
    \hline
\end{tabular}
\caption{\textbf{Definitions of $\widehat{\mathbb{E}}[Q_{i, t} | S_{i, t}]$ for each model class.} $\widehat{\mathbb{E}}[Q_{i, t} | S_{i, t}]$ is the mean of participant model $i$ fitted using data $\{(S_{i, t}, Q_{i, t})\}_{t = 1}^T$. }
\label{table/mean_for_model_classes}
\end{table}

% \alt{ANNA TODO: check if this is being used. I don't think it is so please delete.}
% \begin{table}[!h]
%     \centering
% \begin{tabular}{c|c}
%  \hline
%  \textbf{Model Class} &  $\widehat{\mathbb{E}}[Q_{i, t} | S_{i, t}, Q_{i, t} > 0]$ \\
%  \hline
%     Hurdle (Square Root) & $\sigma_{i,u}^2 + (g(S_{i, t})^Tw_{i, \mu})^2$ \\
%     Zero-Inflated Poisson & $\frac{\exp(g(S_{i, t})^Tw_{i,p})\exp(\exp(g(S_{i, t})^Tw_{i,p}))}{\exp(\exp(g(S_{i, t})^Tw_{i,p})) - 1}$ \\
%     \hline & $\widehat{\text{Var}}[Q_{i, t} | S_{i, t}, Q_{i, t} > 0]$ \\
%     \hline
%     Hurdle (Square Root) &  $g(S_{i, t})^Tw_{i,\mu}^4 + 3\sigma_{i,u}^4 + 6\sigma_{i,u}^2(g(S_{i, t})^Tw_{i,\mu})^2 - \widehat{\mathbb{E}}[R_{i, t} | S_{i, t}, R_{i, t} > 0]^2$ \\
%     Zero-Inflated Poisson & $\widehat{\mathbb{E}}[Q_{i, t} | S_{i, t}, Q_{i, t} > 0] \cdot (1 + \exp(g(S_{i, t})^Tw_{i,p}) - \widehat{\mathbb{E}}[Q_{i, t} | S_{i, t}, Q_{i, t} > 0])$ \\
%     \hline
% \end{tabular}
% \caption{\textbf{Definitions of $\widehat{\mathbb{E}}[Q_{i, t} | S_{i, t}, Q_{i, t} > 0]$} and $\widehat{\text{Var}}[Q_{i, t} | S_{i, t}, Q_{i, t} > 0]$ \textbf{for each model class.} $\widehat{\mathbb{E}}[Q_{i, t} | S_{i, t}, Q_{i, t} > 0]$ and $\widehat{\text{Var}}[Q_{i, t} | S_{i, t}, Q_{i, t} > 0]$ is the mean and variance of the non-zero component of participant model $i$ fitted using data $\{(S_{i, t}, Q_{i, t})\}_{t = 1}^T$.}
% \label{table/non_zero_mean_var_for_model_classes}
% \end{table}

Table~\ref{table/num_model_class} lists the number of model classes for all participants in the ROBAS 3 study that we obtained after the procedure was run.

\begin{table*}[!h]
    \centering
\begin{tabular}{c|c|c}
 \hline
 \textbf{Model Class} & \textbf{Stationary} & \textbf{Non-Stationary} \\
 \hline
    Hurdle (Square Root) & 14 & 12 \\
    Zero-Inflated Poisson & 17 & 19
\end{tabular}
\caption{\textbf{Number of selected model classes for the Stationary and Non-Stationary environments.} Notice that the column numbers sum to 31 corresponding to the 31 ROBAS 3 participants}
\label{table/num_model_class}
\end{table*}
\subsubsection{Imputing Treatment Effect Sizes}
\label{treatment_eff_sizes}
In the previous section, we described how we fit a model for the reward under action 0; this provides baseline effects. In this section, we describe how we constructed the treatment effect sizes to model the reward under action 1. Since this environment is built off of the ROBAS 3 data set which does not have data under intervention (sending an engagement prompt), we use imputed participant-specific treatment effect sizes to model the reward under action 1.

We use three guidelines to design the treatment effect sizes following \cite{a15080255}:
\begin{enumerate}
    \label{effect_size_rationale}
    \item \label{effect_size_rationale:magnitude} For mobile health digital interventions, we expect the treatment effect (magnitude of weight) of actions to be smaller than (or on the order of) the baseline effect of  features (baseline features specified in Section~\ref{baseline_features}).
    \item \label{effect_size_rationale:variance} The variance in treatment effects across participants should be on the order of the variance in the baseline effect of features across participants (i.e., variance in parameters of fitted participant-specific models).
    \item \label{effect_size_rationale:feature_specific} We should have a separate treatment effect size per feature because we believe some features (e.g. day in study) will decrease the effect of sending an engagement prompt as feature values increase and some features (e.g. prior day total OSCB) will increase the effect of sending an engagement prompt as feature values increase. In addition, the treatment effect size on the intercept term should be approximately two times the size of the treatment effect size on other features.
\end{enumerate}

We first construct population-level effect sizes and then use the population effect size to sample unique effect sizes for each participant. Following guidelines \ref{effect_size_rationale:magnitude} and \ref{effect_size_rationale:feature_specific} above, to set the population level effect size per feature, we first take the absolute value of the weights (excluding that for the intercept term) of the participant base models fitted using ROBAS 3 data and then averaged across participants for each feature (e.g., the average absolute value of weight for time of day). We then scale each value by a shrinkage value $\zeta$. Notice that varying the value to $\zeta$ is a way to specify different simulation environment variants. 
% By considering multiple effect size scaling values, we introduce effect size variants to the simulation environment.
We consider two values of $\zeta$ for experiments: $\frac{1}{4}$ and $\frac{1}{8}$ (Section~\ref{procedure_eff_size_scale}) as treatment effects are likely to be much smaller than baseline effects. To construct the population-level effect size on the intercept term, we average the baseline effect size values across features and scale by $2$. We did this procedure to ensure that the effect size on the intercept term is approximately 2 times the effect size of other treatment effect features.

Following guideline \ref{effect_size_rationale:variance} and \ref{effect_size_rationale:feature_specific}, the variance of the normal distribution per feature is found by again taking the absolute value of the weights of the baseline models fitted for each participant, scaling each value by $\zeta$, then taking the empirical variance across participants for each feature. The variance for treatment effect size on the intercept term is the average of empirical variance across features. A more detailed procedure for imputing effect sizes is found in Section~\ref{procedure_eff_size_impute}.
To generate participant-specific effect sizes, for each participant, we draw an effect size vector from a multivariate normal centered at the population effect sizes and a covariance matrix with variance values constructed above along the diagonal. Then we take the absolute value of the effect size vector and depending on the feature, we assign the effect size a positive or negative sign. See Section~\ref{effect_size_sign} for the signs we chose for each feature.

\subsubsection{Treatment Effect Feature Space}
\label{treatment_features}
The treatment effect (advantage) feature space was made after discussions with domain experts on which features are most likely to interact with the intervention (action).

\begin{enumerate}
    \item Time of Day (Morning/Evening) $\in \{0, 1\}$
    \item Prior Day Total OSCB (Normalized) $\in \mathbb{R}$ 
    \item Weekend Indicator (Weekday/Weekend) $\in \{0, 1\}$
    \item Day in Study (Normalized) $\in [-1,1]$
    \item Bias/Intercept Term $\in \mathbb{R}$
\end{enumerate}
The Stationary model uses the state features $h(S_{i, t}) \in \mathbb{R}^4$ that include all features above, except for ``Day in Study". The Non-Stationary model of the base environment uses state $h(S_{i, t}) \in \mathbb{R}^5$ that corresponds to all of the above features.

\subsubsection{Environment Model Including Effect Sizes}
\label{reward_gen_process}
For the zero-inflated Poisson model, we impute treatment effects on both the participant's intent to brush (Bernoulli component) and the participant's brushing duration when they intend to brush (Poisson component). Similarly, for the hurdle models, we impute treatment effects on both whether the participant's brushing duration is zero (Bernoulli component) and the participant's brushing duration when the duration is nonzero. After incorporating treatment effects, OSCB $Q_{i, t}$ under action $A_{i, t}$ in state $S_{i, t}$ is:

$$
Z \sim \text{Bernoulli} \bigg(1 - \mathrm{sigmoid} \big( g(S_{i, t})^\top w_{i,b} - A_{i, t} \cdot \max \big[ \Delta_{i,B}^\top h(S_{i, t}), 0 \big] \big) \bigg)
$$
$$Q_{i, t} =
\begin{cases}
ZY^2, Y \sim \mathcal{N}\big(g(S_{i, t})^Tw_{i,\mu} + A_{i, t} \cdot \max\big[ \Delta_{i,N}^\top h(S_{i, t}), 0 \big], \sigma_u^2\big) & \text{for hurdle square root} \\
ZY, Y \sim \text{Poisson} \big( \exp \big( g(S_{i, t})^\top w_{i,p} + A_{i, t} \cdot \max\big[ \Delta_{i,N}^\top h(S_{i, t}), 0 \big] \big) \big) & \text{for zero-inflated Poisson}
\end{cases}
$$

$\Delta_{i,B}, \Delta_{i,N}$ are participant-specific effect size vectors. $g(S_{i, t})$ is the baseline feature vector as described in Section~\ref{baseline_features}, and $h(S_{i, t})$ is the feature vector that interacts with the effect size specified above.

Notice that our design means the effect size on $Z$ must be non-positive and the effect size on $Y$ component must be non-negative. If this is not the case, then that means in the current context, not sending an engagement prompt will yield a higher OSCB than sending an engagement prompt. %This is unreasonable because the only negative consequences of sending an engagement prompt is the diminishing the responsivity to future messages. 
We ensure that $\max\big[ \Delta_{i,B}^\top h(S_{i, t}), 0 \big]$ and $\max\big[ \Delta_{i,N}^\top h(S_{i, t}), 0 \big]$ are non-negative to prevent the effect size from switching signs and having a negative effect on OSCB.

\subsubsection{Procedure For Imputing Effect Sizes}
\label{procedure_eff_size_impute}
We consider a unique, realistic effect size vector for each participant. We first construct a population-level effect size vector for each of the $Z$ and $Y$ components, $\Delta_B, \Delta_N$ respectively. We then use $\Delta_B, \Delta_N$ to sample participant-specific effect sizes $\Delta_{i,B}, \Delta_{i,N}$.

Recall that for the environment baseline model, we fit a participant-specific model for the OSCB and obtained participant-specific parameters $w_{i,b}, w_{i,p}$ for each environment base model for the $i=1,...,31$ participants in ROBAS 3. We use the fitted parameters to form the population effect sizes as follows.

We use $w_{i,b}^{(d)}, w_{i,p}^{(d)}, w_{i,\mu}^{(d)}$ to denote the $d^{\mathrm{th}}$ dimension of the vector $w_{i,b}, w_{i, p}, w_{i,\mu}$ respectively. Recall that the dimension $d$ corresponds to the treatment effect features in Section~\ref{treatment_features}. We also let $\eta_{i, b}^{(d)} = \frac{1}{8} |w_{i,b}^{(d)}|, \eta_{i, p}^{(d)} = \frac{1}{8} |w_{i,p}^{(d)}|, \eta_{i, \mu}^{(d)} = \frac{1}{8} |w_{i,\mu}^{(d)}|$ denote the transformed parameters. Notice we transform the parameters as a step towards constructing a realistic effect size.  
% \sam{in all of the math below we are acting as if $d=1$ corresponds to the intercept/bias term but we changed the ordering of the features so that the bias/intercept term is $d=5$.  Further the reader has forgotten by now what $d$ even means so need to remind reader of this after we get the ordering straight. }

\textbf{Zero-Inflated Models' Effect Sizes:}
Stationary Environment
\begin{itemize}
    \item $\Delta_B = \big[\frac{1}{N} \sum_{i=1}^N \eta_{i, b}^{(1)}, \frac{1}{N} \sum_{i=1}^N \eta_{i, b}^{(2)}, \frac{1}{N} \sum_{i=1}^N \eta_{i, b}^{(3)}, \frac{1}{3} \sum_{d \in [1 \colon 3]} \frac{1}{N} \sum_{i=1}^N \eta_{i, b}^{(d)}\big]$
    \item $\Delta_N = \big[\frac{1}{N} \sum_{i=1}^N \eta_{i, p}^{(1)}, \frac{1}{N} \sum_{i=1}^N \eta_{i, p}^{(2)}, \frac{1}{N} \sum_{i=1}^N \eta_{i, p}^{(3)}, \frac{1}{3} \sum_{d \in [1 \colon 3]} \frac{1}{N} \sum_{i=1}^N \eta_{i, p}^{(d)}\big]$
\end{itemize}

Non-Stationary Environment
\begin{itemize}
    \item $\Delta_B = \big[\frac{1}{N} \sum_{i=1}^N \eta_{i, b}^{(1)}, 
    \frac{1}{N} \sum_{i=1}^N \eta_{i, b}^{(2)}, 
    \frac{1}{N} \sum_{i=1}^N \eta_{i, b}^{(3)}, 
    \frac{1}{N} \sum_{i=1}^N \eta_{i, b}^{(4)}, 
    \frac{1}{4} \sum_{d \in [1 \colon 4]} \frac{1}{N} \sum_{i=1}^N \eta_{i, b}^{(d)}\big]$
    \item $\Delta_N = \big[\frac{1}{N} \sum_{i=1}^N \eta_{i, p}^{(1)}, 
    \frac{1}{N} \sum_{i=1}^N \eta_{i, p}^{(2)}, 
    \frac{1}{N} \sum_{i=1}^N \eta_{i, p}^{(3)}, 
    \frac{1}{N} \sum_{i=1}^N \eta_{i, p}^{(4)}, 
    \frac{1}{4} \sum_{d \in [1 \colon 4]} \frac{1}{N} \sum_{i=1}^N \eta_{i, p}^{(d)}\big]$
\end{itemize}

\textbf{Hurdle Models' Effect Sizes:}
Stationary Environment
\begin{itemize}
    \item $\Delta_B = \big[\frac{1}{N} \sum_{i=1}^N \eta_{i, b}^{(1)}, \frac{1}{N} \sum_{i=1}^N \eta_{i, b}^{(2)}, \frac{1}{N} \sum_{i=1}^N \eta_{i, b}^{(3)}, \frac{1}{3} \sum_{d \in [1 \colon 3]} \frac{1}{N} \sum_{i=1}^N \eta_{i, b}^{(d)}\big]$
    \item $\Delta_N = \big[\frac{1}{N} \sum_{i=1}^N \eta_{i, \mu}^{(1)}, \frac{1}{N} \sum_{i=1}^N \eta_{i, \mu}^{(2)}, \frac{1}{N} \sum_{i=1}^N \eta_{i, \mu}^{(3)}, \frac{1}{3} \sum_{d \in [1 \colon 3]} \frac{1}{N} \sum_{i=1}^N \eta_{i, \mu}^{(d)}\big]$
\end{itemize}

Non-Stationary Environment
\begin{itemize}
    \item $\Delta_B = \big[\frac{1}{N} \sum_{i=1}^N \eta_{i, b}^{(1)}, 
    \frac{1}{N} \sum_{i=1}^N \eta_{i, b}^{(2)}, 
    \frac{1}{N} \sum_{i=1}^N \eta_{i, b}^{(3)}, 
    \frac{1}{N} \sum_{i=1}^N \eta_{i, b}^{(4)}, 
    \frac{1}{4} \sum_{d \in [1 \colon 4]} \frac{1}{N} \sum_{i=1}^N \eta_{i, b}^{(d)}\big]$
    \item $\Delta_N = \big[\frac{1}{N} \sum_{i=1}^N \eta_{i, \mu}^{(1)}, 
    \frac{1}{N} \sum_{i=1}^N \eta_{i, \mu}^{(2)}, 
    \frac{1}{N} \sum_{i=1}^N \eta_{i, \mu}^{(3)}, 
    \frac{1}{N} \sum_{i=1}^N \eta_{i, \mu}^{(4)}, 
    \frac{1}{4} \sum_{d \in [1 \colon 4]} \frac{1}{N} \sum_{i=1}^N \eta_{i, \mu}^{(d)}\big]$
\end{itemize}

Now to construct the participant-specific effect sizes, we first draw effect sizes for each  participant from a multivariate normal:
$$
\Delta_{i,B}^{(\text{pre})} \sim \text{Multivariate Normal}(\Delta_{B}, \Sigma_{B})
$$
$$
\Delta_{i,N}^{(\text{pre})} \sim \text{Multivariate Normal}(\Delta_{N}, \Sigma_{N})
$$

$\Sigma_{B}, \Sigma_{N}$ are diagonal matrices where the diagonal values are the variance of the fitted parameters over participants for each feature dimension concatenated with the average of those variance values to construct the variance corresponding to the bias term:

\textbf{Zero-Inflated Models' Effect Sizes:}
Stationary Environment
\begin{itemize}
    \item $\Sigma_{B} = \text{diag}\big(\big[\text{Var}(\eta_{i, b}^{(1)}),
    \text{Var}(\eta_{i, b}^{(2)}), \text{Var}(\eta_{i, b}^{(3)}), \frac{1}{3} \sum_{d \in [1 \colon 3]} \text{Var}(\eta_{i,b}^{(d)})\big]\big)$
    \item $\Sigma_N = 
\text{diag}\big(\big[\text{Var}(\eta_{i, p}^{(1)}),
    \text{Var}(\eta_{i, p}^{(2)}), \text{Var}(\eta_{i, p}^{(3)}), \frac{1}{3} \sum_{d \in [1 \colon 3]} \text{Var}(\eta_{i,p}^{(d)})\big]\big)$
\end{itemize}

Non-Stationary Environment
\begin{itemize}
    \item $\Sigma_{B} = \text{diag}\big(\big[\text{Var}(\eta_{i, b}^{(1)}),
    \text{Var}(\eta_{i, b}^{(2)}), 
    \text{Var}(\eta_{i, b}^{(3)}),
    \text{Var}(\eta_{i, b}^{(4)}),
    \frac{1}{4} \sum_{d \in [1 \colon 4]} \text{Var}(\eta_{i, b}^{(d)})\big]\big)$
    \item $\Sigma_N = 
\text{diag}\big(\big[\text{Var}(\eta_{i, p}^{(1)}),
    \text{Var}(\eta_{i, p}^{(2)}), 
    \text{Var}(\eta_{i, p}^{(3)}),
    \text{Var}(\eta_{i, p}^{(4)}),
    \frac{1}{4} \sum_{d \in [1 \colon 4]} \text{Var}(\eta_{i, p}^{(d)})\big]\big)$
\end{itemize}

\textbf{Hurdle Models' Effect Sizes:}
Stationary Environment
\begin{itemize}
    \item $\Sigma_{B} = \text{diag}\big(\big[\text{Var}(\eta_{i, b}^{(1)}),
    \text{Var}(\eta_{i, b}^{(2)}), \text{Var}(\eta_{i, b}^{(3)}), \frac{1}{3} \sum_{d \in [1 \colon 3]} \text{Var}(\eta_{i,b}^{(d)})\big]\big)$
    \item $\Sigma_N = 
\text{diag}\big(\big[\text{Var}(\eta_{i, \mu}^{(1)}),
    \text{Var}(\eta_{i, \mu}^{(2)}), \text{Var}(\eta_{i, \mu}^{(3)}), \frac{1}{3} \sum_{d \in [1 \colon 3]} \text{Var}(\eta_{i,\mu}^{(d)})\big]\big)$
\end{itemize}

Non-Stationary Environment
\begin{itemize}
    \item $\Sigma_{B} = \text{diag}\big(\big[\text{Var}(\eta_{i, b}^{(1)}),
    \text{Var}(\eta_{i, b}^{(2)}), 
    \text{Var}(\eta_{i, b}^{(3)}),
    \text{Var}(\eta_{i, b}^{(4}),
    \frac{1}{4} \sum_{d \in [1 \colon 4]} \text{Var}(\eta_{i, b}^{(d)})\big]\big)$
    \item $\Sigma_N = 
\text{diag}\big(\big[\text{Var}(\eta_{i, \mu}^{(1)}),
    \text{Var}(\eta_{i, \mu}^{(2)}), 
    \text{Var}(\eta_{i, \mu}^{(3)}),
    \text{Var}(\eta_{i, \mu}^{(4)}),
    \frac{1}{4} \sum_{d \in [1 \colon 4]} \text{Var}(\eta_{i, \mu}^{(d)})\big]\big)$
\end{itemize}

We then pass $\Delta_{i,B}^{(\text{pre})}, \Delta_{i,N}^{(\text{pre})}$ into a function $\text{sign}$ that assigns a positive or negative sign to the specific feature.
$$\Delta_{i,B} = \text{sign}(\Delta_{i,B}^{(\text{pre})})$$
$$\Delta_{i,N} = \text{sign}(\Delta_{i,N}^{(\text{pre})})$$

% After following the procedure described above, we found \alt{INSERT VALUES HERE} (values are rounded to the nearest 3 decimal places).

\subsubsection{Checking Quality of Imputed Effect Sizes}
\label{procedure_eff_size_scale}
In this section, we check how reasonable and realistic the imputed effect sizes are for both the stationary and non-stationary variants of the environment. Specifically, we check that the standardized effect sizes we obtain by using the approach in Section~\ref{treatment_eff_sizes} are not unreasonably large or small, according to domain science. Recall that as a part of the imputation procedure, we consider two values to scale the population-level effect size, before sampling unique effect sizes for each participant: a small reasonable effect size (scaled by $\zeta=\frac{1}{8}$) and a less small effect size (scaled by $\zeta=\frac{1}{4}$).
% We describe the procedure for checking the quality of (i.e., how realistic) our imputed effect sizes for both the stationary and non-stationary variants of the environment.

To check that our effect size imputation procedure yields reasonable standardized effect sizes, our procedure is as follows: 1) generate a data set using ROBAS 3 data, 2) fit a linear model to that data set and estimate a population-level standard effect size, and 3) verify the standard effect sizes using domain science.

\paragraph{Step 1. Generate Data set.} We generate a dataset of states and brushing qualities under both action $A_{i, t} = 0$ and $A_{i, t} = 1$:

\begin{enumerate}
    \item We obtain states $S_{i, t}$ for every participant $i$, decision point $t$ from the ROBAS 3 data set.
    \item Recall that we fit a participant-specific environment base model for each participant in ROBAS 3 (Section~\ref{fitting_env_base_models}). We use the participant base models to generate brushing qualities $Q_{i, t}$ for every $S_{i, t}$ under no action ($A_{i, t} = 0$) and under action 1 ($A_{i, t} = 1$). We generate brushing qualities using the following model:
\begin{equation*}
Z \sim \text{Bernoulli} \bigg(1 - \mathrm{sigmoid} \big( g(S_{i, t})^\top w_{i,b} - A_{i, t} \cdot \max \big[\Delta_{i, B}^{(1)}, 0 \big] \big) \bigg)
\end{equation*}
\begin{equation*}
Q_{i, t} =
\begin{cases}
ZY^2, Y \sim \mathcal{N}\big(g(S_{i, t})^Tw_{i,\mu} + A_{i, t} \cdot \max\big[\Delta_{i,N}^{(1)}, 0 \big], \sigma_u^2\big) & \text{for hurdle square root} \\
ZY, Y \sim \text{Poisson} \big( \exp \big( g(S_{i, t})^\top w_{i,p} + A_{i, t} \cdot \max\big[\Delta_{i,N}^{(1)}, 0 \big] \big) \big) & \text{for zero-inflated Poisson}
\end{cases}
\end{equation*}
\end{enumerate}
$\Delta_{i, B}^{(1)}, \Delta_{i,N}^{(1)}$ are the participant-specific effect sizes we constructed on the intercept term in Section~\ref{procedure_eff_size_impute}.

% less adherent participants are equally represented. 
For each participant, we sample with replacement $S_{i, t}, A_{i, t}$ to generate $R_{i, t}$ for 140 decision points. Namely for each participant $i$, we have the following dataset: $D_i = \{(S_{i, t}, A_{i, t}, R_{i, t})\}_{t = 1}^140$.
%this is to make sure that for participants who do not have 140 decision points to get the same tuples for each participant.

% \alt{TODO: document the number of  state action reward tuples per participant and SAR tuples total.}
% \alt{We have a delta b and delta N for each participant and stationry and non-stionary}

\paragraph{Step 2. Obtain Standard Effect Size.} Using the data set generated in Step 1, we calculate the standard effect size:

\begin{enumerate}
    \item Using least squares, we fit $\hat{\theta} = [\hat{\theta_0}, \hat{\theta_1}]$ to the following linear model:
    \begin{equation*}
        R = g(S_{i, t})^T\theta_0 + A_{i, t} \cdot \theta_1
    \end{equation*}
    \item We compute $\hat{\sigma}_{\text{res}}$ the standard deviation of the reward residuals and $\hat{\sigma}_{\text{reward}}$ the standard deviation of the rewards.
    \item We then calculate standardized effect sizes:
    \begin{equation*}
\frac{\hat{\theta_1}}{\hat{\sigma}_{\text{res}}}, \frac{\hat{\theta_1}}{\hat{\sigma}_{\text{reward}}}
    \end{equation*}
\end{enumerate}

% \sam{need to describe what we do with these standardized effect sizes.  That is, we draw reader's attention to rows 7-10 and comment on the effect size being a small effect size.  I forgot why we were aiming around 0.10- 0.15--do you recall why?}
\paragraph{Step 3. Verify Standard Effect Sizes.} Now, with the calculated standard effect sizes, we check that our treatment effect imputation procedure (Section~\ref{procedure_eff_size_impute}) is reasonable. We report values for the imputed effect sizes using the constructed data set in Table~\ref{tab:eff_size_check}. Notice that in rows 7-10, we report the two types of standard effects for both the stationary and non-stationary environments. We verify that effect sizes scaled with $\zeta=\frac{1}{8}$ (small effect sizes) are around 0.1-0.15 and effect sizes scaled with $\zeta=\frac{1}{4}$ (less small effect sizes) are around 0.2-0.25, which is consistent with the domain expert's expected effect sizes.

% \alt{Does anyone know why we were aiming for 0.1-0.15 for small, reasonable effect size and 0.2-0.25 for less small, reasonable effect size? Is this congruent with domain science?}
% Billie says: 0.2 is considered small when testing the effects of standard (macro) intervention components (e.g., coaching sessions) that are expected to impact a distal outcome. When testing micro intervention components (e.g., mobile-based prompts) in relation to a proximal outcome, there are no clear guidelines, but based on my own experience 0.2 would be considered relatively high and 0.1 as small to moderate 

\begin{table}[]
    \centering
    \begin{tabular}{c|c|c}
    Metric & Scaling Value = $\frac{1}{8}$ & Scaling Value = $\frac{1}{4}$ \\
    \toprule
    Stationary $\hat{\theta_1}$ & 8.437 & 13.582 \\
    Non-Stationary $\hat{\theta_1}$ & 8.648 & 18.879 \\
    Stationary $\hat{\sigma}_{\text{res}}$ & 54.542 & 54.788 \\
    Non-Stationary $\hat{\sigma}_{\text{res}}$ & 65.810 & 67.527 \\
    Stationary $\hat{\sigma}_{\text{reward}}$ & 62.152 & 63.546 \\
    Non-Stationary $\hat{\sigma}_{\text{reward}}$ & 75.824 & 78.789\\
    Stationary $\frac{\hat{\theta_1}}{\hat{\sigma}_{\text{res}}}$ & 0.155 & 0.248 \\
    Non-Stationary $\frac{\hat{\theta_1}}{\hat{\sigma}_{\text{res}}}$ & 0.131 & 0.280 \\
    Stationary $\frac{\hat{\theta_1}}{\hat{\sigma}_{\text{reward}}}$ & 0.136 & 0.213\\
    Non-Stationary $\frac{\hat{\theta_1}}{\hat{\sigma}_{\text{reward}}}$ & 0.114 & 0.240 \\
    Number Of Data Tuples & 8680 & 8680
    \end{tabular}
    \caption{Values for checking imputed effect sizes using the constructed data set. Since the residuals within a participant are positively correlated, we believe our estimates of the standard deviation are underestimated. This means our effect sizes are a little bigger than we would expect.}
    \label{tab:eff_size_check}
\end{table}

\subsubsection{Treatment Effect Feature Signs}
\label{effect_size_sign}
Using domain expert knowledge, we assign the following signs for the imputed treatment effect size vector depending on the feature:

\begin{enumerate}
    \item Time of Day (Morning/Evening): Non-negative. We believe participants are more likely to brush if sent an engagement prompt in the evening.
    \item Prior Day Total OSCB (Normalized): Negative. We believe the participants who brush poorly and have a lower prior day total OSCB will be more responsive to an engagement prompt than participants who brush well.
    \item Weekend Indicator (Weekday/Weekend): Non-negative. The effect of getting an engagement prompt on the weekends is higher than on the weekdays because we believe weekends are less structured.
    \item Day in Study (Normalized): Negative. We believe that participants are more responsive to an engagement prompt earlier in the study than at the end.
    \item Bias/Intercept Term: Non-negative.
\end{enumerate}
\subsection{Modeling Delayed Effects of Actions}
\label{sim_env_variants_delayed_effs}
We model delayed effects of actions by shrinking participants' responsiveness to interventions, i.e., their initial treatment effect sizes. We shrink participants' effect sizes by a factor $E \in (0,1)$ when a certain criterion is met. Recall Section~\ref{reward_def} for definitions of $\bar{B}, \bar{A}, b, a_1, a_2$ used in the cost term of the reward definition. Specifically, the criterion if either of the two scenarios holds: (a) $\mathbb{I}[\bar{B}_{i, t} > b]$ (participant brushes well) and $\mathbb{I}[\bar{A}_{i, t} > a_1]$ (participant was sent too many engagement prompts for a healthy brusher), or (b) $\mathbb{I}[\bar{A}_{i, t} > a_2]$ (the participant has been sent too many engagement prompts). 

The first time a participant's criterion has been met, the participant's future effect sizes $\Delta_{i, B}, \Delta_{i, N}$ starting at time $t + 1$ will be scaled down proportionally by $E$ for some $E \in (0,1)$ (the participant is less responsive to treatment). Then after a week, at time $t + 14$, we will check the criterion again. If the criterion is met again, the effect sizes will be further shrunken by a factor of $E$ down to $E^2\cdot\Delta_{i,B}, E^2\cdot\Delta_{i,N}$ starting at time $t + 15$. However, if the criterion is not fulfilled, then the participant recovers their original effect size $\Delta_{i,B}, \Delta_{i,N}$ starting at time $t + 15$. 
This procedure continues until the participant finishes the study. Notice that this means the participant can only have their effect size shrunk at most once a week. This procedure simulates how the participant may experience a reduction in responsiveness, but after a week, if the RL algorithm does not intervene too much, the participant may recover their prior responsivity.
\subsection{Modeling Participants Opening the App}
\label{sim_env_variants_open_app}
We simulate participant app opening behavior to make the simulation environment as close to the study environment as possible. participant app opening behavior is a feature in the state space of the RL algorithm (Section~\ref{alg_state_features}) and is the only way a participant obtains the most recent schedule of actions (Section~\ref{app_open_issue}). 

We define the participant opening the app as having the app in focus (not just in the background). We model a binary 1 if the app is in focus and 0 if not. We describe the procedure for simulating a participant opening their app in the simulation environment. 
For each day in the study, participant $i$ has a probability $p^{\text{app}}$ of opening the app. We sample the participant opening the app from $\text{Bern}(p^{\text{app}})$. 
%$p^{\text{app}}_{i, t}$ has two components: a baseline participant value $p^{\text{base}}_i$ and an effect of intervention $\eta$.
 
\begin{figure}[H]
    \centering
    \includegraphics[width=0.59\textwidth]{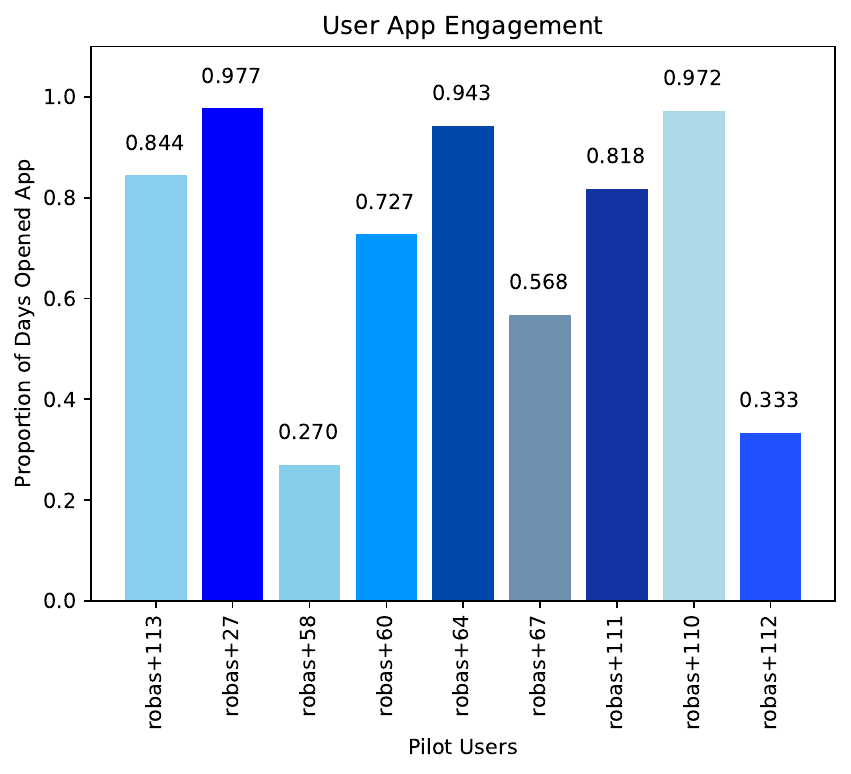}
    \caption{\textbf{App Engagement Data Per Pilot Participant.} Each bar is the number of days a participant opened their app over 35, the total number of days of the pilot study. Reported values are rounded to the nearest 3 decimal places. Notice that the app engagement may appear higher than expected because during the pilot phase, staff implemented a protocol (Section~\ref{staff_protocol}) for contacting participants who did not open their app for 3 days or more and asked them to open their app.}
    \label{fig:app_engage_across_participants}
\end{figure}

\paragraph{App Opening Probability:}
Since ROBAS 3 did not have viable app opening data, we simulated participant app opening as follows. Every participant has the same population-level probability of participants' opening their apps: $p^{\text{app}} = 0.7$. This value is informed by Oralytics pilot data. For each participant in the pilot study, we calculated $p^{\text{app}}_i$, the proportion of days that the participant opened the app during the pilot study (i.e., number of days the participant opened the app divided by 35, the number of days in the pilot study). Then $p^{\text{app}} = \frac{1}{9} \sum_{i = 1}^9 p^{\text{app}}_i$ rounded to the nearest tenth. See Table~\ref{fig:app_engage_across_participants} for $p^{\text{app}}_i$ for each participant in the pilot study. 

\subsection{Simulating App Opening Issue}
\label{sim_env:app_open_issue}
Recall that in Oralytics, a participant only obtains the most recent schedule of actions if they open their app (Section~\ref{app_open_issue}). To simulate this app opening issue in the simulation environment, we have the following procedure:

\begin{enumerate}
    \item For every participant throughout the study, we keep track of: 1) \lq\lq last decision point that participant opened the app" $\in [0, 140]$, 2) \lq\lq prior day app engagement" $\in \{0, 1\}$, and 3) \lq\lq current day app engagement" $\in \{0, 1\}$.
    \item \label{fresh_state_constr} At each decision point $t$, we obtain the participant's \lq\lq prior day app engagement" and use that value (along with other raw features) to construct the most recent algorithm state.
    \begin{itemize}
        \item If $t$ is a \textit{morning} decision point, we sample $O_{i, t} \sim \text{Bern}(p^{\text{app}})$. We set \lq\lq current-day app engagement" (i.e., if the participant opened the app for the current day, the day corresponding to decision points $t$ and $t + 1$) to $O_{i, t}$.  
        %The outcome $O_{i, t}$ represents current day app engagement (i.e., if the participant opened the app for the current day), the day corresponding to decision points $t$ and $t + 1$. 
        If $O_{i, t} = 1$, we update the \lq\lq last decision point that participant opened the app" to $t$. 
        \item If $t$ is a \textit{evening} decision point, we set the \lq\lq prior day app engagement" to $O_{i, t}$ for the next day.
    \end{itemize}
    \item For action-selection corresponding to decision point $t$, if \lq\lq current day app engagement" is 1, then the algorithm selects an action using the fresh state in Step~\ref{fresh_state_constr}. Otherwise, the algorithm selects an action using the stale state as described in Section~\ref{app:modified_rl_features}. Namely, we use the \lq\lq last decision point that the participant opened the app" to get the corresponding stale $\bar{B}$ value for Feature~\ref{modstate:brushing} and set for Feature~\ref{modstate:app} to 0.  %\sam{We already set  \lq\lq prior day app engagement" in step 2...?} %\alt{Hello Susan, thanks for pointing out the confusion. I meant we set the current state feature to 0 and not the tracking variable.}
\end{enumerate}

\clearpage

\bibliography{supp.bib}
\bibliographystyle{apalike}

\end{document}